\documentclass[journal,onecolumn,12pt,draftclsnofoot]{IEEEtran}

\IEEEoverridecommandlockouts
\usepackage{cite}
\usepackage{amsmath,amssymb,amsfonts, amsthm}

\usepackage{authblk}

\usepackage{graphicx}

\usepackage{algorithm}
\usepackage{algpseudocode}
\usepackage{etoolbox,tcolorbox}
\usepackage{pst-plot}
\usepackage{siunitx}
\usepackage{textcomp,mathcomSTEv4}

\theoremstyle{plain}

\usepackage[utf8]{inputenc}

\usepackage{caption}
\usepackage{subcaption}

\usepackage{tikz,pgfplots}
\usetikzlibrary{calc}
\usepgfplotslibrary{fillbetween}
\pgfplotsset{compat=1.10}

\usepackage{pgfplotstable}
\usetikzlibrary{positioning}
\usetikzlibrary{pgfplots.groupplots}
\usetikzlibrary{arrows}
\usetikzlibrary{matrix}
\usepackage{multirow} 
\usepackage{amsthm,enumitem}

\usepackage{xcolor}
\def\BibTeX{{\rm B\kern-.05em{\sc i\kern-.025em b}\kern-.08em
    T\kern-.1667em\lower.7ex\hbox{E}\kern-.125emX}}

\usepackage[compact]{titlesec}
\titlespacing{\subsection}{0pt}{1ex}{0ex}

\newtheorem*{assumption*}{Assumption}
\newtheorem*{remark}{Remark}

\newcommand{\gennorm}{\mathsf{GenNorm}}
\newcommand{\norm}{\mathsf{Norm}}
\newcommand{\laplace}{\mathsf{Laplace}}
\newcommand{\dweibull}{\mathsf{dWeibull}}
\newcommand{\coiii}{\mathsf{CO}_3}
\newcommand{\topK}{ {\mathsf{top}}_K}

\begin{document}

\title{
Communication-Efficient Federated DNN Training: 
Convert, Compress, Correct
}


\author{
Zhong-Jing Chen\textsuperscript{1}
}
\author{
Eduin E. Hernandez\textsuperscript{2}
}
\author{
Yu-Chih Huang\textsuperscript{3}
}
\author{
Stefano Rini\textsuperscript{4}
\thanks{This work was presented in part at the 2022 IEEE International Conference on Communications (ICC) \cite{chen2021dnn}.}
}
\affil{National Yang-Ming Chiao-Tung University (NYCU), Hsinchu City, Taiwan}
\affil{\textit {\{\textsuperscript{1}zhongjing.ee10, \textsuperscript{2}eduin.ee08, \textsuperscript{3}jerryhuang, \textsuperscript{4}stefano.rini\}@nycu.edu.tw}}

\maketitle

\begin{abstract}
This paper introduces $\coiii$ -- an algorithm for communication-efficient federated Deep Neural Network (DNN) training.
$\coiii$  takes its name from three processing applied which reduce the communication load  when transmitting the local DNN gradients from the remote users to the Parameter Server.
%
%
%
%
Namely: 
(i) gradient quantization through floating-point \underline{conversion}, (ii) lossless \underline{compression} of the quantized gradient, and (iii) quantization error \underline{correction}.
We carefully design each of the steps above to assure good training performance under a constraint on the communication rate. 
%
%
%
In particular, in steps (i) and (ii), we adopt the  assumption that DNN gradients are distributed according to a generalized normal distribution, which is validated numerically in the paper.
%
%
For step (iii),  we utilize an error feedback  with memory decay mechanism to correct the quantization error introduced in step (i).
We argue that the memory decay coefficient, similarly to the learning rate, can be optimally tuned to improve convergence. 
%
%
%
A rigorous convergence analysis of the proposed $\coiii$ with SGD is provided.
Moreover, with extensive simulations, we show that $\coiii$ offers improved performance when compared with existing gradient compression schemes in the literature which employ sketching and non-uniform quantization of the local gradients.
%
%
%
\end{abstract}



\section{Introduction}
\label{sec:intro}

%
As the size and complexity of modern-day Deep Neural Networks (DNN) are ever-increasing, high performance is attainable only by training the network over a tremendous amount of data.
In this regime, data centralization is no longer  feasible due to scalability, robustness, and privacy concerns. 
%
%
%
For the reasons above,  federated DNN training has received much attention in the recent literature. 
In many scenarios, the bottleneck in the performance of federated DNN training is often in the communication rate between the remote user and the Parameter Server (PS). 
Motivated by this communication constraint, we investigate 
%
%
the problem of efficiently transmitting DNN model updates from a remote learner to the PS.
More specifically, we consider the Federated Learning (FL)  scenario in which a DNN is trained over datasets present at multiple remote users. 
The communication between the PS and the remote learners is assumed to be an infinite-capacity link, while the link between each learner and the PS is a  finite-capacity link.
We refer to this model as the \emph{rate-limited FL} model.
Rate-limited FL models many scenarios of practical relevance in which a massive number of remote devices participate in the  training of a centralized model through a common communication network -- such as image recognition, speech-to-text models, or recommendation systems. 
Here devices are restricted in uplink transmission rates 
by  power, bandwidth cost, and complexity constraint. 
On the other hand, the latest model can be easily broadcasted from the PS to all nodes.
For this setting, designing communication-efficient schemes for it is a pressing problem to prevent network congestion at the PS. 
To address this issue,  we propose $\coiii$: a novel training/communication scheme for the rate-limited FL setting. 
In $\coiii$, the gradients evaluated at the remote users are first (i) converted to low-resolution floating point representations, then (ii) compressed, losslessly, using the assumption that gradients can be well modelled as 
{
i.i.d. samples}, and then communicated to the PS. 
At the next iteration, the remote users  (iii) correct the quantization error by adding a version of it to the current gradient before (i) is repeated in the next iteration.
A rigorous convergence analysis of the quantized SGD with the proposed $\coiii$ is provided. Moreover, we assess the effectiveness of our algorithm in both a classification task and a teacher-student setting, where the latter involves a randomly initialized first DNN network (the teacher) and a second network (the student) trained to reproduce its output when given a white Gaussian vector as input.
Our simulation results show that this approach provides excellent training performances despite having lower transmission rates between the remote user and the PS than existing arts.



%
\subsection{ Relevant Literature}

In recent years, distributed learning has received considerable attention in the literature \cite{bertsekas2015parallel}. 
%
%
In the following, we shall discuss the communication aspects of FL and distributed training relevant to the development of the paper. 

%
Among various distributed optimization frameworks, 
FL has received particular attention in the recent literature \cite{Shalev-Shwartz2010FL_CE,Wang2018Spars_FL,Alistarh2018Spars_FL,Bernstein2018signSGC,FL_DSGD_binomial,Li2019DP_CEFL}.
FL consists of a central model which is trained   locally at the remote clients by applying Stochastic Gradient Descent (SGD) over a local dataset.
The local gradients are then communicated to the central PS for aggregation into a global model.
A natural constraint in distributed and decentralized optimization is with respect to transmission rates between nodes and its relationship to the overall accuracy \cite{saha2021decentralized,shlezinger2020communication}.
Accordingly, one is interested in devising rate-limited communication schemes that attain high accuracy at a low overall communication payload.
This can be attained through two steps:  (i) dimensionality reduction, and (ii)  quantization and compression. 
The dimensionality-reduction schemes put forth in the literature rely on various sparsification approaches
\cite{Shalev-Shwartz2010FL_CE,Alistarh2018Spars_FL}.
For instance, $\topK$ is a rather aggressive sparsification method that keeps only the coordinates with the largest magnitudes \cite{alistarh2017qsgd,wangni2018gradient}.
Dimensionality-reduction can also be performed on the whole gradient vector as suggested in \cite{gandikota2021vqsgd} through an algorithm referred to as vector Quantized SGD (VQSGD), which leverages the convex hull of particular structured point sets to produce an unbiased gradient estimate that has a bounded variance, thus reducing the communication cost while ensuring convergence guarantees.
In \cite{salehkalaibar2022lossy}, the authors propose a choice of distortion which promotes sparse gradient quantization, conceptually generalizing $\topK$.

For quantization and compression approaches, the gradients are digitized through quantization, either scalar-wise \cite{Konecny2016Fl_CE,seide2014onebitSGD,salehkalaibar2022lossy} or vector-wise \cite{gandikota2021vqsgd}.
From an implementation-oriented perspective, \cite{sun2019hybrid} studies the effect of gradient quantization when constrained to a \emph{sign-exponent-mantissa} representation.
%
%
%

%
%

After quantization, enabled by the statistical model obtained via extensive simulations that gradients in DNN training with SGD follow i.i.d. generalized normal ($\gennorm$) distribution, lossless compression can be applied to further reduce the communication rate toward the PS in a preliminary version of the present work \cite{chen2021dnn}. 
In the scheme of \cite{rothchild2020countsketch}, each client performs local compression to the local stochastic gradient by count sketch via a common sketching operator. 
In \cite{fangcheng2020tinyscript}, the authors introduced a non-uniform quantization algorithm, TINYSCRIPT, to compress the activations and gradients of a DNN.

When  gradients are compressed, it has been shown that error correction,  or error feedback, can greatly improve performance \cite{karimireddy2019error}. 
Error feedback for $1$-bit quantization was originally considered in \cite{seide2014onebitSGD}.
In \cite{stich2018sparsified}, error feedback is applied to gradient compression in a more general manner than \cite{seide2014onebitSGD}.  
%



\subsection{Contributions}
In this paper, we consider the problem of efficient gradient compression for rate-limited federated DNN training. 
%
The main contributions of the paper can be summarized as follows:

\begin{enumerate}[ labelwidth=!, labelindent=-95pt,label=\alph*)]
\item {\bf Sec. \ref{sec:Gradient modelling} --} \underline{Validation of the $\gennorm$ assumption:}
A fundamental assumption of our algorithm design is the observation that the DNN gradient can be well approximated as i.i.d. generalized normal ($\gennorm$) samples. 
In Sec. \ref{sec:Gradient modelling} we numerically validate this assumption over various DNN training examples.
It is worth emphasizing that through validating the $\gennorm$ assumption, we are able to build an accurate statistical model for gradients in DNN training. This statistical model is of practical importance in its own right.
\item {\bf Sec. \ref{sec:proposed approach} --} \underline{Design of $\coiii$:}
Using the $\gennorm$ assumption as a stepping stone, we propose $\coiii$ as an efficient  algorithm for the distributed training of DNN models over communication-constrained FL scenario.
\item {\bf Sec. \ref{sec:Theoritical proof} --} \underline{Convergence analysis of SGD with $\coiii$:}
We provide a rigorous proof of the convergence of SGD with $\coiii$ under the $\gennorm$ assumption.

\end{enumerate}

Let us comment on Sec. \ref{sec:Gradient modelling} and \ref{sec:proposed approach} in more details.

\begin{enumerate}[ labelwidth=!, labelindent=-5pt,label=a.\roman*)]
\item \underline{GenNorm modelling:}
To the best of our knowledge, a good statistical model for modelling gradients in DNN training is currently lacking.
 We argue that one can effectively model such gradients as i.i.d. $\gennorm$ variables. 
We also  argue that  the gradient distribution approaches the normal  distribution as the depth of the network increases and as the epoch number increases. Additionally,  we contend that the kurtosis of the gradient distribution provides a useful measure of the concentration of gradient around zero.
 
 \item \underline{Numerical validation:}
 We use statistical methods to validate the $\gennorm$ assumption for three DNN architectures in the image classification task and teacher-student setting, namely DenseNet\cite{huang2017densely},  ResNet\cite{he2016identity}, and  NASNet\cite{zoph2018learning},
 across both layers and training epochs.

\item \underline{Error feedback:}
In FL, error feedback is an effective mechanism to compensate the quantization error at the remote users. 
 We argue, in Sec. \ref{subsec:error_feedback}, that the $\gennorm$ assumption remains valid when error feedback is employed in distributed SGD.
\end{enumerate}

Having numerically validated our assumption on the gradient distribution, we then move to the design of a communication-efficient algorithm for distributed DNN training. 
The proposed scheme, which we term $\coiii$,  comprises of the following three gradient processing steps:  

\begin{enumerate}[ labelwidth=!, labelindent=-5pt,label=b.\roman*)]
  \item  \underline{Floating point (fp) {\bf conversion}:} As a quantization mechanism, we consider fp conversion due to its compatibility with GPU gradient processing.
 As in \cite{sun2019hybrid}, we choose the exponent bias to minimize the $L_2$ loss between the original and quantized gradient. 
 
  \item \underline{Lossless gradient {\bf compression}:} 
  {Leveraging the $\gennorm$ modeling of the local gradients, we further compress the quantized gradients using lossless compression in the form of Huffman coding.} 
  \item \underline{Error {\bf  correction}:} The quantization error is stored in one iteration and corrected in the next.
  We further employ a memory decay coefficient  which prevents the accumulation of stale gradients.
\end{enumerate}
%

\smallskip
\noindent
{\bf Notation.}
Lowercase boldface letters (e.g., $\zv$) are used for tensors, 
uppercase letters for random variables (e.g. $X$), and calligraphic uppercase for  sets (e.g. $\Acal$) .
Given the set $\Acal$, $|\Acal|$ indicates the cardinality of the set. 
We also adopt the short-hands  $[m:n] \triangleq \{m, \ldots, n\}$
and  $[n] \triangleq \{1, \ldots, n\}$. 
Both subscripts and superscripts letters (e.g. $g_t$ and $g^{(u)}$) indicate the iteration index and the user index for a tensor, respectively.
%
The superscript $\Tsf$ (e.g. $g^{\Tsf}$) denotes the transpose of the tensor.
The all zero vector is indicated as $\zerov$.
$\mathbb{E}[X]$ represents the expected value of random variable $X$.
Finally, $\Fbb_2$ is the binary field.

\section{System Model}
\label{sec:System Model}

In many FL scenarios of practical relevance, the communication from the remote users to the PS is severely constrained in terms of transmission rate.
For this reason, in the following,  we consider the approach of \cite{shlezinger2020communication,chen2021dnn,salehkalaibar2022lossy} and investigate the accuracy/payload trade-off of distributed optimization. 
%
%
In Sec. \ref{sec:DNN training} we specialized the general setting of \cite{shlezinger2020communication,chen2021dnn} to the specific case of federated DNN training.

\subsection{Federated Optimization}
\label{sec:Distributed Optimization}

Consider the scenario with $U$ remote users, each possessing a local dataset
\ea{
\Dc^{(u)} = \left\{\lb \dv_{k}^{\lb u\rb},v_k^{\lb u\rb}\rb\right\}_{k\in\left[\left|\Dc^{(u)}\right|\right]},
}
where $\Dc^{(u)}$  includes
$\left|\Dc^{(u)}\right|$
pairs, each comprising a data point 
$\dv_{k}^{\lb u\rb}$
and the label $v_{k}^{(u)}$
for $u \in [U]$.  
Users collaborate with the PS to minimize the loss function $\Lcal$ as evaluated across all the local datasets and over the choice of the model $\wv \in \Rbb^d$,
that is 
\ea{
\Lcal(\wv) =  \f 1 {|\Dcal|} \sum_{u \in [U]} \sum_{k\in\left[\left|\Dcal^{(u)}\right|\right]} \Lcal(\wv;\dv^{(u)}_{k},v^{(u)}_{k}),
\label{eq:loss}
}
where $\Dcal$ is defined as $\Dcal = \cup_{u \in  } \Dcal^{(u)}$.
%
For the \emph{loss function} $\Lcal$ in the LHS of \eqref{eq:loss}, we assume that there exists a unique minimizer $\wv^*$, which we referred to as the \emph{optimal model}.
%

A common approach for numerically determining this unique minimizer, $\wv^*$, is through the iterative application of (synchronous) SGD.
In the SGD algorithm, the model parameter $\wv$ is updated at each iteration $t$, 
by taking a step toward the negative direction of the stochastic gradient vector,   that is 
%
\ea{
\wv_{t+1}=\wv_{t}-\eta_t  \gv_t,
\label{eq:SGD}
}
for $t \in [T]$, a choice of initial model $\wv_{0}$, and where  $\gv_t$ is the stochastic gradient of $\Lcal(\cdot)$ evaluated in $\wv_{t}$, 
that is
$\Ebb\lsb \gv_t\rsb=  \nabla\Lcal\lb\wv_t,\Dcal^{(u)}\rb$.
Finally, $\eta_t$ in \eqref{eq:SGD} is  an iteration-dependent step size, called \emph{learning rate}.

In the FL setting, the SGD iterations  are distributed among $U$ users and is orchestrated by PS as follows:
(i) each user $u \in[U]$ receives the current model estimate, $\wv_t$ of the optimal model $\wv^*$ over the infinite capacity link from the PS.
The user $u \in [U]$ then (ii) accesses its  local dataset $\Dc^{(u)}$ and computes the local stochastic gradient $\gv_t^{(u)}$.
%
%
Finally (iii) each node communicates the gradient estimate $\gv_{t}^{(u)}$ to the PS which then computes the term $\gv_t$  
as
\ea{
\gv_t=\f{1}{U}\sum_{u \in[U]} \gv_t^{(u)},
\label{eq:aggregate}
}
and uses $\gv_t$ to update the model estimate.
We refer to the above FL training algorithm as  \emph{federate averaging} (FedAvg)  \cite{mcmahan2016federated}.

\subsection{Rate-limited Federated Optimization}
\label{sec:Rate-limited distributed training}

%
%

In the rate-limited distributed training scenario \cite{shlezinger2020communication,chen2021dnn,salehkalaibar2022lossy}, communication between each user and the PS takes place over a noiseless channel with finite capacity $\Rsf$. 
On the other hand, communication between the PS and remote users is unconstrained. 
%
Without loss of generality, the user is therefore required to convert the $d$-dimensional gradient vector $\gv_{nt}$ to a $\Rsf$ binary vector. 
A general three-step scheme to address the finite transmission rate $\Rsf$ between each user and the PS can be defined as follows:
%


\smallskip
\noindent
{\bf (i) Pre-processing:}
Assume that at time $t$ the local gradient $\gv_{t}^{(u)}$ becomes available at the remote user $u$. 
This value is fed into a function $f:\mathbb{R}^d\rightarrow\mathbb{R}^d$ to form $\vv_{t}^{(u)}=f(\gv_{1}^{(u)},\ldots,\gv_{t}^{(u)})$. Here, the function $f$ is introduced to allow pre-processing before quantization. In our work, $f$ is employed as a correction mechanism for the quantization error. 
We note that when $f(\mathbf{x})=\mathbf{x}$, i.e., the identity function, then no pre-processing takes place. 

\smallskip
\noindent
{\bf (ii) Quantization:} the term $\vv_{t}^{(u)}$  is  quantized via a quantizer $Q:\mathbb{R}\rightarrow \mathcal{X}$ to form the representative $\vhv_t^{(u)}=Q(\vv_{t}^{(u)})$, where $\mathcal{X}$ is the collection of representatives, i.e., quantization levels.
The mapping $Q$ is meant as a stochastic mapping, so that non-subtractive dither quantization can be implemented.

\smallskip
\noindent
{\bf (iii) Lossless compression:}
Following quantization, the quantized gradients are further compressed through the mapping $h:\mathcal{X}\rightarrow \mathbb{F}_2^{*}$ to form a codeword $\mathbf{b}_t^{(u)}=h(\vhv_t^{(u)})$. Here, we allow $h$ to be a variable-length coding scheme; hence, the range is $\mathbb{F}_2^{*}$, where $\mathbb{F}_2$ is the binary field.  
This compression step is lossless, that is, the mapping $h$ is invertible: The role of this mapping is to remove the statistical redundancy inherent in the local gradients, thus reducing the amount of bits to be transmitted to the PS. 
Finally, let $r_t^{(u)}$ be the length of $\mathbf{b}_t^{(u)}$.

\smallskip
\noindent
{\bf (iv) Aggregation:}
Upon receiving the transmission from  the user $u \in [U]$ at time $t$, the PS reconstructs the gradient estimate $\ghv_t^{(u)}$.
After reconstruction, 
the estimated gradient is used to update the network as
\ea{
\whv_t= \whv_{t-1}-  \f{\eta_t}{U}\sum_{u \in[U]} \ghv_t^{(u)}.
\label{eq:aggregate hat}
}

The above steps are repeated for all $t \in [T]$. At time $t=T$
the model $\whv_T$ is declared as the approximate value of the optimal model, $\wv^*$.

\subsection{Gradient Assumptions}
A meaningful engineering approach to the design of the steps in Sec.\ref{sec:Rate-limited distributed training} can be established only under some regularity assumption on the local gradients observed through training and across initialization. 
In this paper we shall adopt the following assumption:
%
%
Let us assume that  the local gradients are distributed i.i.d. according to $\mathbb{P}_{\Gv_t}$ at each user, that is $\gv_t^{(u)} \sim \mathbb{P}_{G_t}^d$ for all $u \in [U]$. 
To be more precise, we assume the gradients in one layer are independent of but non-identically distributed to those in another layer. i.e., $\mathbb{P}_{G_t}^d$ is in fact the product of $L$ distributions.
Also, note that across users the gradients are assumed to be independent and identically distributed.
This follows from the fact that the local gradient at different users is an unbiased estimate of the underlying true gradient.
We notice that the mean field theory has provided a partial validation of this i.i.d. assumption. 
In a series of extremely interesting papers
\cite{mei2018mean,araujo2019mean,nguyen2020rigorous,fang2021modeling}, it has been shown that, in various regimes, DNN weights in a given layer become  indistinguishable as  the number of SGD steps grows large.
%
For instance, \cite{araujo2019mean} considers a DNN with a fixed number of layers and any width  trained to minimize  the square loss over an i.i.d. dataset.
In this scenario, the authors show that  training through SGD is well-approximated  by continuous dynamics expressed through certain non-linear partial differential equations.
As a result of this approximation, the weights can be well-approximated through \emph{distributional dynamics} \cite{mei2019mean} which shows that the per-layer DNN weights are indistinguishable. 
%


\subsection{Problem formulation}
\label{sec:Problem formulation}

Under the assumption that a gradient distribution can be properly defined, we can then define the expectation of the  performance of a scheme in the framework of Sec. \ref{sec:Rate-limited distributed training}.
Remembering that $r_t^{(u)}$ is the length of $\mathbf{b}_t^{(u)}$, the transmitted binary string, we define the expected length of $u\in[U]$ at $t\in[T]$ as
${R_t^{(u)} = \mathbb{E}_{f,Q,h} [r_t^{(u)} ],}$
where the expectation is taken w.r.t. $\prod_{t' \in [t]}\Pbb_{\Gv_{t'}}$ the product gradient distribution until time $t$.
The {\em communication overhead} of a certain choice  of functions $(f,Q,h)$ as the  \underline{sum of expected lengths} conveyed over the up-link channel  over the training, that is 
	\begin{equation}
	\label{eqn:Overhead}
	\Rsf = \sum_{t \in [T]} \sum_{u \in [U]} R_t^{(u)}.
	\end{equation}
	%
%
Using the definition in \eqref{eqn:Overhead}, we finally  come to the definition of the accuracy/overhead trade-off 
as 
%
\ea{
\mathsf{P}_c(\Rsf) = 
 1-\mathsf{P}_e(\wv_T, \Dcal_{U}) ,
\label{eq:Lsf}
}
where  $\Dcal_{U} = \cup_{u \in [U]} \Dcal^{(u)}$, $\mathsf{P}_e(\wv_T, \Dcal_{U})$ denotes the probability of prediction error for the network $\wv_T$.
%
%
In other words, $\Psf_T(\Rsf)$ is the accuracy that one can attain in $T$ iterations when the total communication payload is $\Rsf$ under $(f,Q,h)$.
Note  that the minimization is over the pre-processing, quantization, and lossless compression operations.
\subsection{DNN training}
\label{sec:DNN training}
While we have so far considered a general distributed optimization problem,  in the remainder of the paper we shall consider the particular case of DNN training. 
%
%
More specifically, we consider the training for the CIFAR-10 dataset classification task using the following three architectures: (i) DenseNet121, (ii) ResNet50V2, and (iii) NASNetMobile. 
For each architecture, the training is performed using SGD optimizer with a constant $\eta=0.01$ learning rate in \eqref{eq:aggregate hat}.
The rest of the configurations of the parameters and hyperparameters used for the training are specified in Tab. \ref{tab:DNN parameters}.

We also validate our assumption and scheme with the teacher-student setting \cite{hinton2015distilling}. This setting consists of two neural networks, where the teacher model is fixed and the student model is trained using the output of the teacher model, given the same input dataset for both models. This setup enables the weight parameters of the student model to learn the distribution of those of the teacher model. In our training setup, we utilize the same CNN architecture for both the teacher and student models. The inputs to both models are drawn i.i.d from the standard normal distribution, and their weights are initialized according to the initialization. Validation has been conducted under the teacher-student Gaussian setting, which serves to bolster the $\gennorm$ assumption with regard to generalization.

\begin{table}
	\footnotesize
	\centering
	\vspace{0.04in}\caption{Parameters and hyperparameters used for the training of the DNN models.}
	\label{tab:DNN parameters}
	\begin{tabular}{|c|c|}
		\hline
		Dataset & CIFAR-10 \\ \hline
        Training Samples & \num{50000} \\ \hline
        Test Samples & \num{10000} \\\hline
        Optimizer & SGD \\ \hline
        Learning Rate & \num{0.01} \\ \hline
        Momentum & 0 \\ \hline
        Loss & Categorical Cross Entropy \\ \hline
        Epochs & 150 \\ \hline
        Mini-Batch Sizes & 64 \\ \hline
	\end{tabular}
	\vspace{-0.5cm}
\end{table}

During each batch-iteration, the gradients of the trainable parameters are accumulated on a temporal memory on a per-layer basis with the intention on averaging them along the epoch.
At the end of the epoch, the gradients are saved and the temporal memory is freed.
This process is repeated until the last epoch for the gradient analysis provided in the next subsections. The code for the gradient modeling and analysis is available at  https://github.com/Chen-Zhong-Jing/CO3\_algorithm.
As these are very deep networks as specified in Tab. \ref{tab:total network parameters}, we will limit the scope to three layers in each of the architectures: One 2-dimensional convolution layer located in the upper, middle, and lower sections of the networks.
Tab. \ref{tab:network parameters} details the number of trainable weight parameters for these chosen layers.
%

\begin{table}
    \begin{minipage}{.49\linewidth}
    	\footnotesize
        \caption{} \label{tab:total network parameters}
        \caption{Total number of layers, weight parameters, and trainable weight parameters belonging to each architecture.} 
        \centering
        \begin{tabular}{|c|c|c|c|}
        \hline{Architectures}
        & Layers & Total Params & Train Params \\
        \hline{DenseNet121}
        & 121 & \num{7047754} & \num{6964106} \\
        \hline{ResNet50V2}
        & 50 & \num{23585290} & \num{23539850} \\
        \hline{NASNetMobile}
        & - & \num{4280286} & \num{4243548} \\
        \hline
        \end{tabular}
    \end{minipage}
    \hspace{0.3cm}
    \begin{minipage}{.49\linewidth}
        \footnotesize
        \caption{}  \label{tab:network parameters}
        \caption{Number of trainable weight parameters for the chosen layers of each architecture.\\} 

        \centering
        \begin{tabular}{|c|c|c|c|}
        \hline{Architectures}
        & Upper & Middle & Lower\\
        \hline{ResNet50V2}
        & \num{4096} & \num{32768} & \num{524288} \\
        \hline{DenseNet121}
        & \num{9408} & \num{16384} & \num{65536} \\
        \hline{NASNetMobile}
        & \num{3872} & \num{30976} & \num{185856} \\
        \hline
        \end{tabular}
    \end{minipage}
	\vspace{-0.5cm}
\end{table}

\subsection{Further comments}

Before delving further into the paper, let us add a few brief comments on some aspects of the problem formulation in  Sec. \ref{sec:Problem formulation} which are not considered in the paper. 

\smallskip
\noindent
{\bf Asynchronous/delayed gradients:}
In the paper, we only consider the case in which training occurs \emph{simultaneously and synchronously} at all remote users. 
This is an idealized assumption that rarely holds in practical scenarios. 
Although asynchronous training and delayed gradients are very pressing and interesting problem, it is not considered in the following.

\smallskip
\noindent
{\bf Across-layer correlation:}
In the following, we compress the gradient distribution as independent across iterations.
Although correlation of the gradients across layers exists, we do not consider such correlation. 
%
For instance, in \cite{araujo2019mean},  it is shown that the per-layer distribution is conditionally dependent only on the weights in the previous layer. 
For simplicity, in the following, we do not consider this dependency. 
The design of a version of the proposed approach taking advantage of this correlation is left for future research. 
%

\smallskip
\noindent
{\bf Lossless compression:}
In this paper, we consider the \emph{lossless compression} scenarios in which the PS is interested in the exact reconstruction of the quantized gradients $\vhv_t^{(u)}$ from $\mathbf{b}_t^{(u)}$.
Our choice is generally dictated by the fact that a precise understanding of the effect of the distortion criteria used for compression on the learning performance is unclear. 
For instance, $\topK$ sparsification  \cite{shi2019understanding} suggests that an appropriate choice of  distortion should take into account the gradient magnitude.
This is in contrast with the compression error introduced by the classic mean squared error (MSE) criteria which is commonly used in practical lossless compression algorithms.
%
%
%
The authors of  \cite{salehkalaibar2022lossy} consider a weighted norm approach that bridges the compression through $\topK$  and that of MSE criteria. 
The intuition between this choice distortion is the fact that the gradients with a larger magnitude are more relevant to the DNN training.

\smallskip
\noindent
{\bf Universal compression:}
In the paper, we consider distribution-based compression. When the underlying gradient distribution
is unknown, one can employ Lempel-Ziv coding \cite{ziv1978compression}, which is asymptotically optimal in terms of the expected length, even when the gradients are generated by a rather general (and yet unknown) source \cite{kontoyiannis1997second}.
Despite such wide applicability, the performance of such a universal source coding scheme is generally not acceptable in the short to medium-source length regime. 
In contrast, in the presence of knowledge about $\mathbb{P}_{\tilde{G}_t}$, optimal lossless compression can be easily achieved by Huffman coding \cite{cover_book}. 
%
%
Additionally, universal approaches have a generally higher implementation complexity and memory occupation which might not be amenable to low-cost and fast implementation. 
%
%
%





\section{Gradient Modelling for DNN Training}
\label{sec:Gradient modelling}

As a first contribution of the paper, we wish to argue that the DNN gradient distribution $\mathbb{P}_{G_t}$, if exists, can be well modeled, in each layer, as an i.i.d. $\gennorm$ distribution. It should be emphasized that we do not make any assumptions regarding the knowledge of the DNN gradient distribution $\mathbb{P}_{G_t}$. Instead, we rely on the $\gennorm$ distribution family to provide an approximation for $\mathbb{P}_{G_t}$. By using the $\gennorm$ modelling, we are able to estimate the parameters of the distribution using the gradient samples obtained during training, which allows us to apply compression schemes without relying on any prior knowledge about the true distribution of the gradients.
The $\gennorm$ parameters are assumed to change across layers and across iterations but all variables are otherwise independent.

%
%

In its parametrization through a location $\mu$, scale $\al$, and shape $\beta$ parameters, the pdf of the $\gennorm$ distribution is obtained as  
\ea{
& X \sim \gennorm(x,\mu, \al, \be)  \implies 
f_X(x)= \f {\be}{2\al \Gamma(1/\be)} \exp \lcb 
-  \lb \f  { \labs x-\mu\rabs} {\al} \rb^{\be}\rcb,  
\label{eq:gennorm}
}
%
The mean, variance, and kurtosis of $X \sim \gennorm(x,\mu, \al, \be)$ have the following expressions:
%
\begin{equation}
\mathrm{Mean}(X)  =\mu, \quad
\mathrm{Var}(X)  = \frac{\alpha^2\Gamma(3/\beta)}{\Gamma(1/\beta)},\quad
\kappa\triangleq \mathrm{Kurt}(X)  = \frac{\Gamma(5/\beta)\Gamma(1/\beta)}{\Gamma(3/\beta)^2}.
\label{eq:kurtosis}
\end{equation}

$\gennorm$ is a family of distributions that subsumes Laplace ($\be=1$) and Normal $(\be=2)$ distributions as special cases.
%
%
When the shape parameter $\be < 2$, the distribution is leptokurtic and has a heavier tail than the normal distribution.

With the consideration above, we now state the working assumption that shall guide the design of the proposed scheme, $\coiii$, in Sec. \ref{sec:proposed approach}.

\begin{assumption*}{\bf GenNorm DNN gradients:}
For each layer and for each epoch, the DNN gradients are distributed according to the $\gennorm$ distribution in \eqref{eq:gennorm}.
\end{assumption*}
For brevity, we refer to the above assumption as the $\gennorm$ assumption.
With respect to the validity of the $\gennorm$ assumption, we notice that recently the authors of \cite{isik2022information} have argued that gradients posses a $\laplace$ distribution, partially supporting our assumption. 

In the remainder of the section, we shall motivate the $\gennorm$ assumption from a statistical perspective and further describe the dependency of the parameters on the layer depth and iteration number. 

\begin{remark}[\emph{Error-Feedback}]
In this section we shall consider the case in which no gradient pre-processing is performed, that is $f(\xv)=\xv$. 
In $\coiii$ we will consider the case in which $f$ is an error feedback mechanism similar to that in \cite{Alistarh2018Spars_FL}.
In other words, the quantization error is accumulated and added as a correction term when quantizing the gradient in the new iteration. 
In order to improve the flow of the paper, the validation of the $\gennorm$ assumption with error feedback is
presented in  Sec. \ref{subsec:error_feedback}.
\end{remark}


 

\subsection{Statistical Validation}
\label{subsec:stat_valid}
\begin{figure}
    \centering
    \begin{tikzpicture}[thick, scale=0.9]
    \definecolor{mycolor1}{rgb}{0.00000,0.44706,0.74118}%
    \definecolor{mycolor2}{rgb}{0.63529,0.07843,0.18431}%
    \definecolor{mycolor3}{rgb}{0.00000,0.49804,0.00000}%
    \begin{groupplot}[
        group style={
            group name=my plots,
            group size=3 by 2,
            xlabels at=edge bottom,
            xticklabels at=edge bottom,
            vertical sep=0pt,
            horizontal sep=5pt
        },
        height=5cm,
        width=5.5cm,
        xmax=0.006,
        xmin=-0.006,
        ymax=1100,
        ymin=0,
        xlabel={x},
        xtick={-0.005, -0.0025, 0, 0.0025, 0.005},
        xmajorgrids,
        ymajorgrids
    ]
    \nextgroupplot[title={Epoch 2}, yticklabels=\empty, ylabel={ResNet50V2}]
        \coordinate (top) at (axis cs:0,\pgfkeysvalueof{/pgfplots/ymin}) -- (axis cs:0,\pgfkeysvalueof{/pgfplots/ymax});
        \addplot+[ybar interval,mark=no,fill=blue!120,draw=blue, opacity=0.5]
                [restrict x to domain=-0.005:0.005]
                table[x index=0, y index=1]{./Data/ResNet50V2_top_layer_e2_hist_without_quantization.txt};
        \addplot [fill=red!120,draw=red, opacity=0.4]
                [restrict x to domain=-0.005:0.005]
                table[x index=0, y index=1]{./Data/ResNet50V2_top_layer_e2_pdf_without_quantization.txt};
        \addplot [fill=green!120,draw=green, opacity=0.35]
                [restrict x to domain=-0.005:0.005]
                table[x index=0, y index=2]{./Data/ResNet50V2_top_layer_e2_pdf_without_quantization.txt};
    \nextgroupplot[title={Epoch 50}, yticklabels=\empty]
        \coordinate (top) at (axis cs:0,\pgfkeysvalueof{/pgfplots/ymin}) -- (axis cs:0,\pgfkeysvalueof{/pgfplots/ymax});
        \addplot+[ybar interval,mark=no,fill=blue!120,draw=blue, opacity=0.5]
                [restrict x to domain=-0.005:0.005]
                table[x index=0, y index=1]{./Data/ResNet50V2_top_layer_e50_hist_without_quantization.txt};
        \addplot [fill=red!120,draw=red, opacity=0.4]
                [restrict x to domain=-0.005:0.005]
                table[x index=0, y index=1]{./Data/ResNet50V2_top_layer_e50_pdf_without_quantization.txt};
        \addplot [fill=green!120,draw=green, opacity=0.35]
                [restrict x to domain=-0.005:0.005]
                table[x index=0, y index=2]{./Data/ResNet50V2_top_layer_e50_pdf_without_quantization.txt};
    \nextgroupplot[title={Epoch 100}, yticklabels=\empty]
        \addplot+[ybar interval,mark=no,fill=blue!120,draw=blue, opacity=0.5]
                [restrict x to domain=-0.005:0.005]
                table[x index=0, y index=1]{./Data/ResNet50V2_top_layer_e100_hist_without_quantization.txt};
        \addplot [fill=red!120,draw=red, opacity=0.4]
                [restrict x to domain=-0.005:0.005]
                table[x index=0, y index=1]{./Data/ResNet50V2_top_layer_e100_pdf_without_quantization.txt};
        \addplot [fill=green!120,draw=green, opacity=0.35]
                [restrict x to domain=-0.005:0.005]
                table[x index=0, y index=2]{./Data/ResNet50V2_top_layer_e100_pdf_without_quantization.txt};
    \nextgroupplot[yticklabels=\empty, ylabel={NASNetMobile}]
        \addplot+[ybar interval,mark=no,fill=blue!120,draw=blue, opacity=0.5]
                [restrict x to domain=-0.005:0.005]
                table[x index=0, y index=1]{./Data/NASNetMobile_top_layer_e2_hist_without_quantization.txt};
        \addplot [fill=red!120,draw=red, opacity=0.4]
                [restrict x to domain=-0.005:0.005]
                table[x index=0, y index=1]{./Data/NASNetMobile_top_layer_e2_pdf_without_quantization.txt};
        \addplot [fill=green!120,draw=green, opacity=0.35]
                [restrict x to domain=-0.005:0.005]
                table[x index=0, y index=2]{./Data/NASNetMobile_top_layer_e2_pdf_without_quantization.txt};
    \nextgroupplot[yticklabels=\empty]
        \addplot+[ybar interval,mark=no,fill=blue!120,draw=blue, opacity=0.5]
                [restrict x to domain=-0.005:0.005]
                table[x index=0, y index=1]{./Data/NASNetMobile_top_layer_e50_hist_without_quantization.txt};
        \addplot [fill=red!120,draw=red, opacity=0.4]
                [restrict x to domain=-0.005:0.005]
                table[x index=0, y index=1]{./Data/NASNetMobile_top_layer_e50_pdf_without_quantization.txt};
        \addplot [fill=green!120,draw=green, opacity=0.35]
                [restrict x to domain=-0.005:0.005]
                table[x index=0, y index=2]{./Data/NASNetMobile_top_layer_e50_pdf_without_quantization.txt};
    \nextgroupplot[yticklabels=\empty]
        \addplot+[ybar interval,mark=no,fill=blue!120,draw=blue, opacity=0.5]
                [restrict x to domain=-0.005:0.005]
                table[x index=0, y index=1]{./Data/NASNetMobile_top_layer_e100_hist_without_quantization.txt};
        \addplot [fill=red!120,draw=red, opacity=0.4]
                [restrict x to domain=-0.005:0.005]
                table[x index=0, y index=1]{./Data/NASNetMobile_top_layer_e100_pdf_without_quantization.txt};
        \addplot [fill=green!120,draw=green, opacity=0.35]
                [restrict x to domain=-0.005:0.005]
                table[x index=0, y index=2]{./Data/NASNetMobile_top_layer_e100_pdf_without_quantization.txt};
    \end{groupplot}
\end{tikzpicture}
    \caption{Histogram of gradient (blue) with PDF of $\gennorm$ (red) and $\norm$ (green) of top layer for epoch 2, 50, and 100 in two different network.}
    \label{fig:histogram}
	\vspace{-0.5cm}
\end{figure}
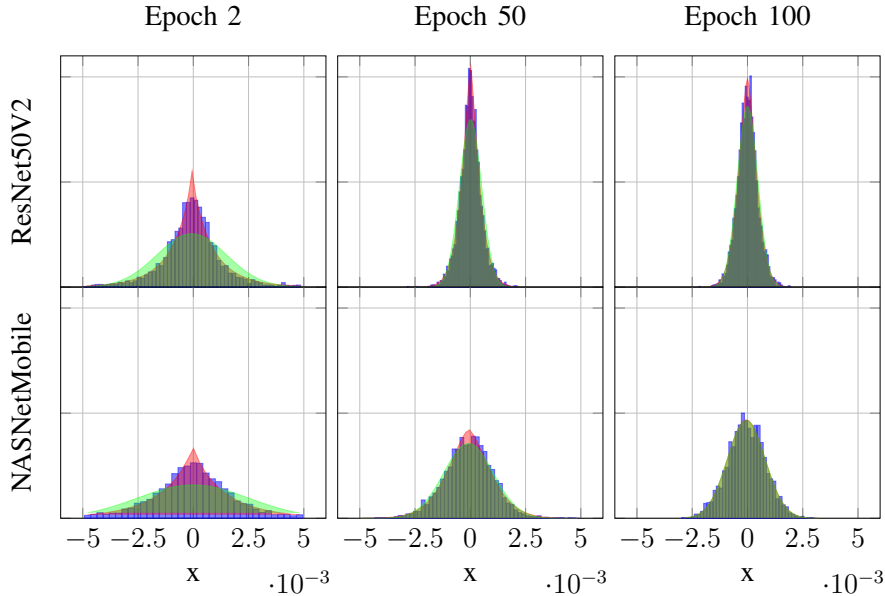

Let us begin by visually inspecting the gradient histogram for the networks in Sec. \ref{sec:DNN training}, as depicted in 
Fig. \ref{fig:histogram}.
In this figure, we plot (i) the sample distribution, (ii) the $\gennorm$ fitting, and (iii) the normal distribution ($\norm$) fitting for ResNet50V2 and NASNetMobile across three epochs: $2$, $50$, and $100$.
We observe that in the earlier epochs, the gradient histogram is closer to the $\gennorm$ distribution in that the sample distribution is (i) more concentrated in zero, and  (ii) it contains heavier tails than the $\norm$ distribution. 
As the training continues, the  variance of the gradient distribution gradually reduces and approaches the $\norm$ distribution.
For instance, the gradients from ResNet50V2 seem to converge to the $\norm$ distribution slower than NASNetMobile.

Fig. \ref{fig:histogram} only provides a qualitative depiction of the $\gennorm$ assumption. 
A quantitative depiction is provided in Fig. \ref{fig:full_w2}, where we plot the 1D $W_2$ Wasserstein distance \cite{villani_book}, defined as 
%
$W_2(X,Y)=\left(\int_{0}^{1}|F^{-1}_{X}(z)-F^{-1}_{Y}(z)|dz\right)^{1/2}, $
between the modelled distributions: (i) $\gennorm$,
(ii) $\norm$, (iii) $\laplace$, and (iv) double Weibull ($\dweibull$) and the gradient samples
 for DenseNet121 and NasNetMobile as a function of the epoch number in CIFAR-10 classfication task.
%
Note that Wasserstein distance is a distance metric measuring the distance between two probability distributions and is recently very popular for many machine learning applications due to its many nice properties.
From Fig. \ref{fig:full_w2} left, both $\dweibull$ and $\gennorm$ distribution provide a closer fitting with gradient samples than the $\norm$ and $\laplace$.
Notice that $\laplace$ and $\norm$ distribution maintain a constant separation between their errors in comparison to $\gennorm$ since the first two only require optimization of the scale parameter, $\alpha$, in comparison to $\gennorm$ which also requires tuning of $\beta$.
The same goes for $\dweibull$.
With the same number of parameters, $\gennorm$ provides slightly lower loss than that of $\dweibull$.
From Fig. \ref{fig:full_w2} right, the $\laplace$ distribution is outperformed by the $\norm$ after the first few epochs, but $\gennorm$ and $\dweibull$ still maintain the advantage, showing their flexibility in adapting the tails to better fit the distribution while $\gennorm$, once again, shows slightly better fitting than all distributions.

In Fig. \ref{fig:nasnet_densenet_cifar10_vanilla_w2}, we employ the same validation for the teacher-student scheme, which is mentioned in Sec.\ref{sec:DNN training}. For the upper layer of DenseNet121, the $\gennorm$ and $\norm$ distribution both provide the best fitting among the four distribution families. However, the $\laplace$, $\dweibull$ and $\gennorm$ are closer to the gradient distribution of lower layer in DenseNet121. Overall, the $\gennorm$ distribution consistently offers a better approximation of the gradient distribution for different layers in various CNN architectures.
%
%

\subsection{Distribution parameters}
\label{subsec:distribution_parameter}
\begin{figure}
    \centering
    \begin{subfigure}{.4\textwidth}
        \centering
        \begin{tikzpicture}[scale = 0.8]
    \definecolor{mycolor1}{rgb}{0.00000,0.44706,0.74118}%
    \definecolor{mycolor2}{rgb}{0.63529,0.07843,0.18431}%
    \definecolor{mycolor3}{rgb}{0.00000,0.49804,0.00000}%
    \begin{axis}[
    ymin=-5,
    ymax=51,
    xmin=0,
    xmax=100,
    xlabel={Epoch},
    ylabel={Excess Kurtosis},
    grid=both]
        \addplot [draw=mycolor1, line width=1.5pt, mark=diamond, mark options={solid, mycolor1}, mark repeat={10}, smooth]
            table[x index=0, y index=1]{./Data/DenseNet121_excess_kur.txt};
            \addlegendentry{Upper}
        \addplot [draw=mycolor2, line width=1.5pt, mark=square, mark options={solid, mycolor2}, mark repeat={10}, smooth]
            table[x index=0, y index=2]{./Data/DenseNet121_excess_kur.txt};
            \addlegendentry{Middle}
        \addplot [draw=mycolor3, line width=1.5pt, mark=o, mark options={solid, mycolor3}, mark repeat={10}, smooth]
            table[x index=0, y index=3]{./Data/DenseNet121_excess_kur.txt};
            \addlegendentry{Lower}
    \end{axis}
\end{tikzpicture}
        \vspace{-1.0cm}
        \label{fig:densenet_kurtosis}
    \end{subfigure}
    \begin{subfigure}{.4\textwidth}
        \centering
        \begin{tikzpicture}[scale = 0.8]
    \definecolor{mycolor1}{rgb}{0.00000,0.44706,0.74118}%
    \definecolor{mycolor2}{rgb}{0.63529,0.07843,0.18431}%
    \definecolor{mycolor3}{rgb}{0.00000,0.49804,0.00000}%
    \begin{axis}[
    ymin=-5,
    ymax=58,
    xmin=0,
    xmax=100,
    xlabel={Epoch},
    ylabel={Excess Kurtosis},
    grid=both]
        \addplot [draw=mycolor1, line width=1.5pt, mark=diamond, mark options={solid, mycolor1}, mark repeat={10}, smooth]
            table[x index=0, y index=1]{./Data/nasnetmobile_excess_kurtosis.txt};
            \addlegendentry{Upper}
        \addplot [draw=mycolor2, line width=1.5pt, mark=square, mark options={solid, mycolor2}, mark repeat={10}, smooth]
            table[x index=0, y index=2]{./Data/nasnetmobile_excess_kurtosis.txt};
            \addlegendentry{Middle}
        \addplot [draw=mycolor3, line width=1.5pt, mark=o, mark options={solid, mycolor3}, mark repeat={10}, smooth]
            table[x index=0, y index=3]{./Data/nasnetmobile_excess_kurtosis.txt};
            \addlegendentry{Lower}
    \end{axis}
\end{tikzpicture}
        \vspace{-1.0cm}
        \label{fig:nasnet_kurtosis}
    \end{subfigure}
    \vspace{-0.5cm}
    \caption{Excess kurtosis of each convolution layer for DenseNet121 (left) and NASNetMobile (right).}
    \label{fig:kurtosis}
	\vspace{-0.5cm}
\end{figure}
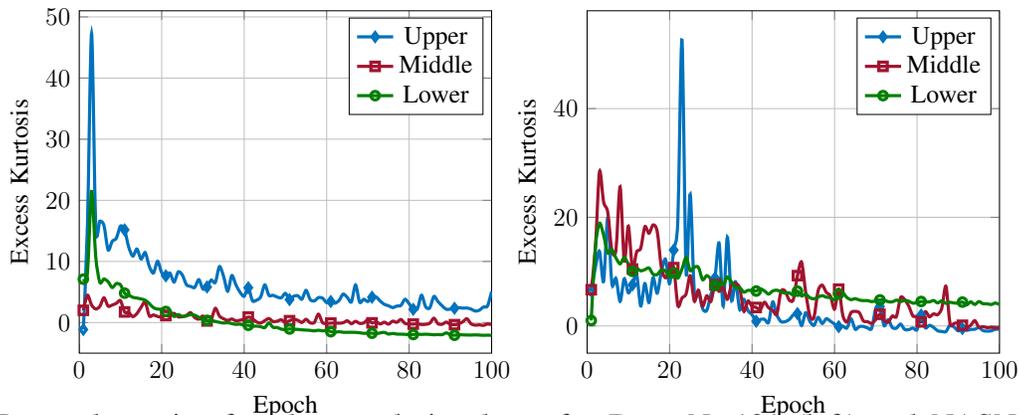

The mean and variance of the sample gradient distribution are provided in Tab. \ref{tab:network mean and variance}, together with the respective  confidence interval. 
Another important aspect of the $\gennorm$ is that it highlights the role of the kurtosis in describing the behaviour of the gradients, as in \eqref{eq:kurtosis}, the kurtosis depends only on the parameter $\be$.
In Fig. \ref{fig:kurtosis}, we plot the excess kurtosis (i.e., the kurtosis minus 3). 
%
Observation of the excess kurtosis for the upper layers in Fig. \ref{fig:kurtosis} provides us further insight into previous behavior from Fig. \ref{fig:full_w2}: for  DenseNet121, the excess kurtosis suggests the distribution requires longer tails than $\norm$ while for the NasNetMobile later epochs suggests the excess kurtosis approximates the $\norm$ distribution, hence why $\laplace$ had a lower loss than $\norm$ in Fig. \ref{fig:full_w2} left and why the opposite happened in Fig. \ref{fig:full_w2} right.
Results in these figures again indicate the $\gennorm$ modeling tending towards the $\norm$ with further epochs (excess kurtosis evolving from positive to near 0).
%
%



\begin{figure}
    \centering
    \begin{subfigure}{.4\textwidth}
        \centering
        \begin{tikzpicture}[scale = 0.8]
    \definecolor{mycolor1}{rgb}{0.00000,0.44706,0.74118}%
    \definecolor{mycolor2}{rgb}{0.63529,0.07843,0.18431}%
   \definecolor{mycolor3}{rgb}{0.00000,0.49804,0.00000}%
   \definecolor{mycolor4}{rgb}{0.60000,0.19608,0.80000}
    \begin{axis}[
    ymin=0,
    xmin=0,
    xmax=150,
    xlabel={Epoch},
    ylabel={Wasserstein Distance},
   grid=both]
        \addplot [draw=mycolor1, line width=1.5pt, mark=diamond, mark options={solid, mycolor1}, mark repeat={2},each nth point=5, smooth]
            table[x index=0, y index=1]{./Data/densenet121_upper_w2.txt};
            \addlegendentry{Normal}
        \addplot [draw=mycolor2, line width=1.5pt, mark=square, mark options={solid, mycolor2}, mark repeat={2},each nth point=5, smooth]
            table[x index=0, y index=2]{./Data/densenet121_upper_w2.txt};
            \addlegendentry{Laplace}
        \addplot [draw=mycolor3, line width=1.5pt, mark=diamond, mark options={solid, mycolor3}, mark repeat={2},each nth point=5, smooth]
            table[x index=0, y index=3]{./Data/densenet121_upper_w2.txt};
            \addlegendentry{dWeibull}
        \addplot [draw=mycolor4, line width=1.5pt, mark=diamond, mark options={solid, mycolor4}, mark repeat={2},each nth point=5, smooth]
            table[x index=0, y index=4]{./Data/densenet121_upper_w2.txt};
            \addlegendentry{GenNorm}
    \end{axis}
\end{tikzpicture}
        \vspace{-1.0cm}
        \label{fig:densenet_full_w2}
    \end{subfigure}
    \begin{subfigure}{.4\textwidth}
        \centering
        \begin{tikzpicture}[scale = 0.8]
    \definecolor{mycolor1}{rgb}{0.00000,0.44706,0.74118}%
    \definecolor{mycolor2}{rgb}{0.63529,0.07843,0.18431}%
    \definecolor{mycolor3}{rgb}{0.00000,0.49804,0.00000}%
    \definecolor{mycolor4}{rgb}{0.60000,0.19608,0.80000}
    \begin{axis}[
    ymin=0,
    xmin=0,
    xmax=150,
    xlabel={Epoch},
    ylabel={Wasserstein Distance},
   grid=both]
        \addplot [draw=mycolor1, line width=1.5pt, mark=diamond, mark options={solid, mycolor1}, mark repeat={2},each nth point=5, smooth]
            table[x index=0, y index=1]{./Data/nasnet_upper_w2.txt};
            \addlegendentry{Normal}
        \addplot [draw=mycolor2, line width=1.5pt, mark=square, mark options={solid, mycolor2}, mark repeat={2},each nth point=5, smooth]
            table[x index=0, y index=2]{./Data/nasnet_upper_w2.txt};
            \addlegendentry{Laplace}
        \addplot [draw=mycolor3, line width=1.5pt, mark=diamond, mark options={solid, mycolor3}, mark repeat={2},each nth point=5, smooth]
            table[x index=0, y index=3]{./Data/nasnet_upper_w2.txt};
            \addlegendentry{dWeibull}
        \addplot [draw=mycolor4, line width=1.5pt, mark=diamond, mark options={solid, mycolor4}, mark repeat={2},each nth point=5, smooth]
            table[x index=0, y index=4]{./Data/nasnet_upper_w2.txt};
            \addlegendentry{GenNorm}
    \end{axis}
\end{tikzpicture}
        \vspace{-1.0cm}
        \label{fig:nasnet_full_w2}
    \end{subfigure}
    \vspace{-0.5cm}
    \caption{$W_2$ distance between the empirical  CDF and best-fit CDF for gradient in upper layer of DenseNet121 (left) and NASNetMobile (right) for various gradient assumptions.}
    \label{fig:full_w2}
	\vspace{-0.3cm}
\end{figure}
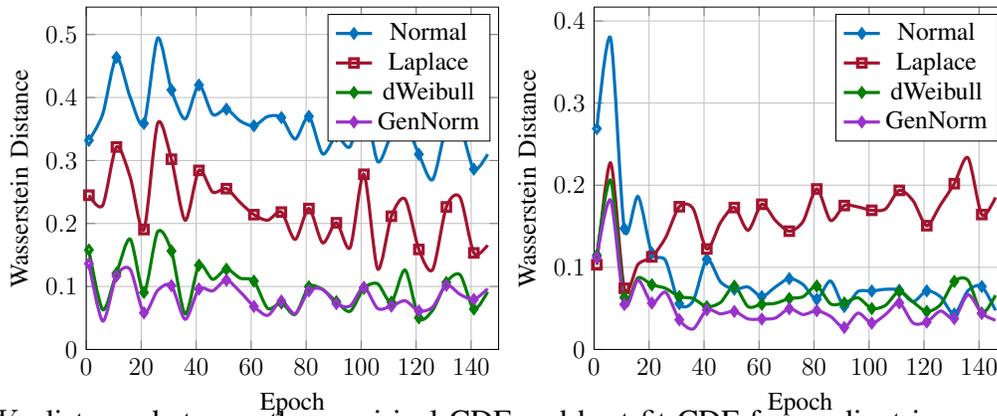

\begin{figure*}[t]
    \centering
	\begin{tikzpicture}[thick, scale=0.85]
    \definecolor{mycolor1}{rgb}{0.00000,0.44706,0.74118}%
    \definecolor{mycolor2}{rgb}{0.63529,0.07843,0.18431}%
    \definecolor{mycolor3}{rgb}{0.00000,0.49804,0.00000}%
    \definecolor{mycolor4}{rgb}{0.60000,0.19608,0.80000}%
    \begin{groupplot}[
        group style={
            group name=my plots,
            group size=2 by 3,
            xlabels at=edge bottom,
            xticklabels at=edge bottom,
            vertical sep=0pt
        },
        height=3.5cm, 
        width=9cm, 
        xmin=0,
        xlabel=Epoch,
        grid=both
    ]
    \nextgroupplot[title=NASNetMobile,xmax=50,
    ylabel=Upper]
        \coordinate (top) at (axis cs:1,\pgfkeysvalueof{/pgfplots/ymax});
        \addplot [draw=mycolor1, line width=1.5pt, mark=diamond, mark options={solid, mycolor1}, mark repeat={2},each nth point=2,smooth]
            table[x index=0, y index=1]{./Data/teacher_student/vanilla/NASNetMobile_upper_w2.txt};
            \label{plots:plot1_nasnet_w2}
        \addplot [draw=mycolor2, line width=1.5pt, mark=diamond, mark options={solid, mycolor2}, mark repeat={2},each nth point=2, smooth]
            table[x index=0, y index=2]{./Data/teacher_student/vanilla/NASNetMobile_upper_w2.txt};
            \label{plots:plot2_nasnet_w2}
        \addplot [draw=mycolor3, line width=1.5pt, mark=square, mark options={solid, mycolor3}, mark repeat={2},each nth point=2, smooth]
            table[x index=0, y index=3]{./Data/teacher_student/vanilla/NASNetMobile_upper_w2.txt};
            \label{plots:plot3_nasnet_w2}
        \addplot [draw=mycolor4, line width=1.5pt, mark=square, mark options={solid, mycolor4}, mark repeat={2},each nth point=2, smooth]
            table[x index=0, y index=4]{./Data/teacher_student/vanilla/NASNetMobile_upper_w2.txt};
            \label{plots:plot4_nasnet_w2}

    \nextgroupplot[title=DenseNet121 ,xmax=50]
        \coordinate (top) at (axis cs:1,\pgfkeysvalueof{/pgfplots/ymax});
        \addplot [draw=mycolor1, line width=1.5pt, mark=diamond, mark options={solid, mycolor1}, mark repeat={2},each nth point=2,smooth]
            table[x index=0, y index=1]{./Data/teacher_student/vanilla/DenseNet121_upper_w2.txt};
            \label{plots:plot1_nasnet_w2}
        \addplot [draw=mycolor2, line width=1.5pt, mark=diamond, mark options={solid, mycolor2}, mark repeat={2},each nth point=2, smooth]
            table[x index=0, y index=2]{./Data/teacher_student/vanilla/DenseNet121_upper_w2.txt};
            \label{plots:plot2_nasnet_w2}
        \addplot [draw=mycolor3, line width=1.5pt, mark=square, mark options={solid, mycolor3}, mark repeat={2},each nth point=2, smooth]
            table[x index=0, y index=3]{./Data/teacher_student/vanilla/DenseNet121_upper_w2.txt};
            \label{plots:plot3_nasnet_w2}
        \addplot [draw=mycolor4, line width=1.5pt, mark=square, mark options={solid, mycolor4}, mark repeat={2},each nth point=2, smooth]
            table[x index=0, y index=4]{./Data/teacher_student/vanilla/DenseNet121_upper_w2.txt};
            \label{plots:plot4_nasnet_w2}
            
    \nextgroupplot[ylabel style={align=center},xmax=50,
    ylabel=Middle]
        \addplot [draw=mycolor1, line width=1.5pt, mark=diamond, mark options={solid, mycolor1}, mark repeat={2},each nth point=2, smooth]
            table[x index=0, y index=1]{./Data/teacher_student/vanilla/NASNetMobile_middle_w2.txt};
        \addplot [draw=mycolor2, line width=1.5pt, mark=diamond, mark options={solid, mycolor2}, mark repeat={2},each nth point=2, smooth]
            table[x index=0, y index=2]{./Data/teacher_student/vanilla/NASNetMobile_middle_w2.txt};
        \addplot [draw=mycolor3, line width=1.5pt, mark=square, mark options={solid, mycolor3}, mark repeat={2},each nth point=2, smooth]
            table[x index=0, y index=3]{./Data/teacher_student/vanilla/NASNetMobile_middle_w2.txt};
        \addplot [draw=mycolor4, line width=1.5pt, mark=square, mark
        options={solid, mycolor4}, mark repeat={2}, each nth point=2, smooth]
            table[x index=0, y index=4]{./Data/teacher_student/vanilla/NASNetMobile_middle_w2.txt};

    \nextgroupplot[ylabel style={align=center},xmax=50]
        \addplot [draw=mycolor1, line width=1.5pt, mark=diamond, mark options={solid, mycolor1}, mark repeat={2},each nth point=2, smooth]
            table[x index=0, y index=1]{./Data/teacher_student/vanilla/DenseNet121_middle_w2.txt};
        \addplot [draw=mycolor2, line width=1.5pt, mark=diamond, mark options={solid, mycolor2}, mark repeat={2},each nth point=2, smooth]
            table[x index=0, y index=2]{./Data/teacher_student/vanilla/DenseNet121_middle_w2.txt};
        \addplot [draw=mycolor3, line width=1.5pt, mark=square, mark options={solid, mycolor3}, mark repeat={2},each nth point=2, smooth]
            table[x index=0, y index=3]{./Data/teacher_student/vanilla/DenseNet121_middle_w2.txt};
        \addplot [draw=mycolor4, line width=1.5pt, mark=square, mark options={solid, mycolor4}, mark repeat={2},each nth point=2, smooth]
            table[x index=0, y index=4]{./Data/teacher_student/vanilla/DenseNet121_middle_w2.txt};
    
    \nextgroupplot[ylabel style={align=center},xmax=50,
    ylabel=Lower]
        \addplot [draw=mycolor1, line width=1.5pt, mark=diamond, mark options={solid, mycolor1}, mark repeat={2},each nth point=2, smooth]
            table[x index=0, y index=1]{./Data/teacher_student/vanilla/NASNetMobile_lower_w2.txt};
        \addplot [draw=mycolor2, line width=1.5pt, mark=diamond, mark options={solid, mycolor2}, mark repeat={2},each nth point=2, smooth]
            table[x index=0, y index=2]{./Data/teacher_student/vanilla/NASNetMobile_lower_w2.txt};
        \addplot [draw=mycolor3, line width=1.5pt, mark=square, mark options={solid, mycolor3}, mark repeat={2},each nth point=2, smooth]
            table[x index=0, y index=3]{./Data/teacher_student/vanilla/NASNetMobile_lower_w2.txt};
        \addplot [draw=mycolor4, line width=1.5pt, mark=square, mark options={solid, mycolor4}, mark repeat={2},each nth point=2, smooth]
            table[x index=0, y index=4]{./Data/teacher_student/vanilla/NASNetMobile_lower_w2.txt};

    \nextgroupplot[ylabel style={align=center},xmax=50]
        \addplot [draw=mycolor1, line width=1.5pt, mark=diamond, mark options={solid, mycolor1}, mark repeat={2},each nth point=2, smooth]
            table[x index=0, y index=1]{./Data/teacher_student/vanilla/DenseNet121_lower_w2.txt};
        \addplot [draw=mycolor2, line width=1.5pt, mark=diamond, mark options={solid, mycolor2}, mark repeat={2},each nth point=2, smooth]
            table[x index=0, y index=2]{./Data/teacher_student/vanilla/DenseNet121_lower_w2.txt};
        \addplot [draw=mycolor3, line width=1.5pt, mark=square, mark options={solid, mycolor3}, mark repeat={2},each nth point=2, smooth]
            table[x index=0, y index=3]{./Data/teacher_student/vanilla/DenseNet121_lower_w2.txt};
        \addplot [draw=mycolor4, line width=1.5pt, mark=square, mark options={solid, mycolor4}, mark repeat={2},each nth point=2, smooth]
            table[x index=0, y index=4]{./Data/teacher_student/vanilla/DenseNet121_lower_w2.txt};
            
    \coordinate (bot) at (axis cs:1,\pgfkeysvalueof{/pgfplots/ymin});
    \end{groupplot}

    \path (top|-current bounding box.north)--
          coordinate(legendpos)
          (bot|-current bounding box.north);
    \matrix[
        matrix of nodes,
        anchor=south,
        draw,
        inner sep=0.2em,
        draw
      ]at([yshift=1ex,xshift=-5ex]legendpos)
      {
        \ref{plots:plot1_nasnet_w2}& normal &[3pt]
        \ref{plots:plot2_nasnet_w2}& Laplace &[3pt]
        \ref{plots:plot3_nasnet_w2}& dWeibull &[3pt]
        \ref{plots:plot4_nasnet_w2}& $\gennorm$ &[3pt]\\};
\end{tikzpicture}
    \vspace{-0.5cm}
    \caption{$W_2$ distance between the empirical  CDF and best-fit CDF for gradient in each layer of NASNetMobile (left) and DenseNet121 (right) in the teacher-student setting.}
    \label{fig:nasnet_densenet_cifar10_vanilla_w2}
    \vspace{-0.5cm}
\end{figure*}
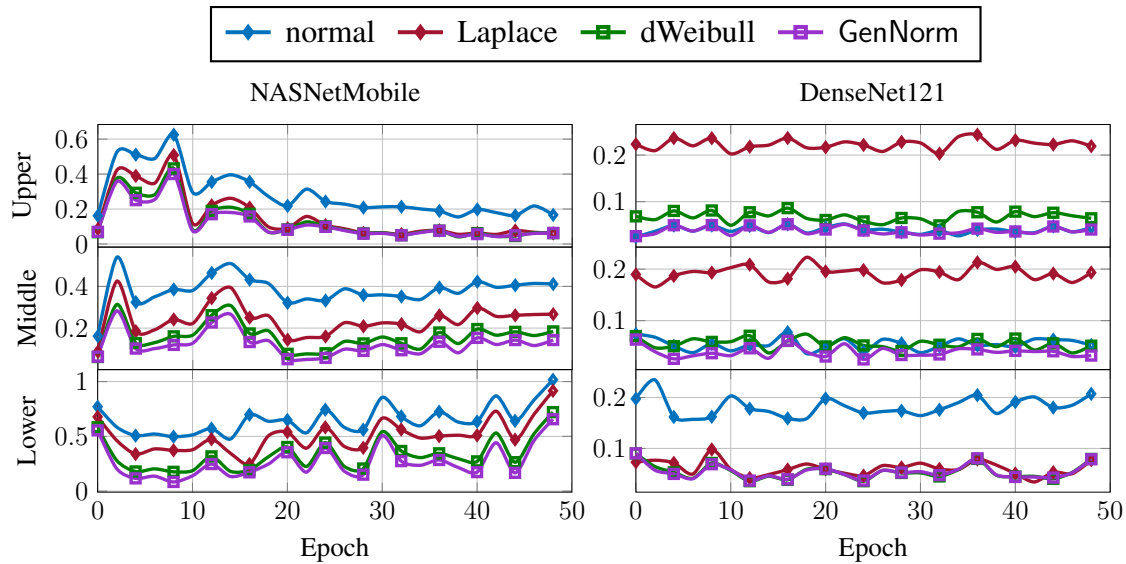



\begin{table*}[t]
    \footnotesize
    \vspace{0.04in}\caption{The mean and variance of NASNetMobile's gradient at different epochs} 
    \label{tab:network mean and variance}
    \centering
    \begin{tabular}{|c|c|c|c|c|}
    \hline
    
    \hline{Layers}
    & & Epoch 2 & Epoch 50 & Epoch 100\\
    \hline
    \multirow{2}{*}{Upper}
    & mean 
    & \num{-1.72e-05} $\pm$ \num{1.12e-07}
    & \num{-8.38e-05} $\pm$ \num{1.90e-08}
    & \num{-5.53e-05} $\pm$ \num{2.13e-08}
    \\ \cline{2-5}
    & variance 
    & \num{9.58e-05}$\pm$ \num{8.03e-08}
    & \num{1.86e-05}$\pm$ \num{3.82e-10}
    & \num{1.07e-05}$\pm$ \num{1.82e-11}
    \\ \hline
     \multirow{2}{*}{Middle}
    & mean 
    & \num{6.66e-06}$\pm$ \num{9.53e-09}
    & \num{1.10e-05}$\pm$ \num{8.57e-09}
    & \num{3.68e-06}$\pm$ \num{1.92e-09} 
    \\ \cline{2-5}
    & variance 
    & \num{4.96e-06}$\pm$ \num{1.19e-10}
    & \num{3.56e-06}$\pm$ \num{7.66e-11}
    & \num{1.93e-06}$\pm$ \num{7.31e-13} 
    \\ \hline
     \multirow{2}{*}{Lower}
    & mean 
    & \num{8.97e-05}$\pm$\num{4.40e-09}
    & \num{4.85e-06}$\pm$\num{3.88e-10}
    & \num{5.10e-07}$\pm$\num{3.13e-11}
    \\ \cline{2-5}
    & variance 
    & \num{4.74e-07 }$\pm$ \num{4.99e-14}
    & \num{2.24e-07}$\pm$ \num{2.52e-15} 
    & \num{1.96e-07}$\pm$ \num{6.30e-16} 
    \\ \hline
    \end{tabular}
	\vspace{-0.5cm}
\end{table*}

\section{Proposed Approach: $\coiii$}
\label{sec:proposed approach}

In this section, we propose a novel algorithm for communication-efficient distributed DNN training, named $\coiii$. 
This algorithm implements the operations of Sec. \ref{sec:Rate-limited distributed training}: 
(i) pre-processing is error correction with a memory decay, that is $f$ that accumulates a version of the quantization error of the previous epochs;
(ii) quantization is chosen as fp conversion, optimized over the exponent bias. Finally, compression is Huffman coding using the $\gennorm$ assumption over the quantized gradients. 
In Sec.~\ref{subsec:CO}, we introduce the algorithm in its general form.  
In Sec.~\ref{subsec:CO_DNN}, we specialize the proposed $\coiii$ to DNN models and detail the operations in the proposed algorithm. 
Since the algorithm requires knowledge of the underlying distribution of quantization outputs, in Sec.~\ref{subsec:error_feedback}, we return to the approach in Sec. \ref{sec:Gradient modelling} and argue that the $\gennorm$ assumption is again valid even in the presence of error feedback. 
Finally, in Sec.~\ref{subsec:error_mag}, we investigate the magnitude of the error term and that of the gradient during the learning process to determine the weighting of the memory in the proposed $\coiii$ algorithm.


\subsection{$\coiii$ algorithm}
\label{subsec:CO}
In the following, we propose a novel algorithm, named $\coiii$, which falls within the general framework described in Sec. \ref{sec:Rate-limited distributed training}.
More precisely, $\coiii$ considers the following gradient processing steps (i)  the quantization, $Q$, is chosen as the fp conversion, (ii) compression, $h$, is chosen as element-wise Huffman coding, and (iii) an $f$ that performs error correction with a memory decay of $\gamma$.

\smallskip
Next, let us describe each of the steps above in further detail: 

\smallskip
\noindent
{\bf (i) fp conversion:} the local gradient $\gv_t^{(u)}$ is converted into the IEEE fp representation $\sf[sgn, exp, mant]$ \cite{ieee754ref} with ${\sf sgn} =1$ bit for the sign, $\sf exp$ bits for the exponent and $\sf mant$ bits for the mantissa. Furthermore, at each time $t$, we introduce a bias $b_t$ on the exponent so as to minimize the expected loss between the closest quantization representative and $\gv_t^{(u)}$, that is
\ea{
b_t = \argmin_b  \Ebb \lsb |  c_{\sf sgn} \cdot c_{\sf mant}  \cdot 2^{ c_{\sf exp} +b} - G_t  |^2\rsb,
\label{eq:exp bias}
}
where $b \in \Rbb$ and the expectation is over the gradient distribution $\Pbb_{\Gv_t}$, and
$c_{\sf sgn}\in\{-1,1\}$ determined by ${\sf sgn}$, $c_{\sf exp}\in\mathbb{Z}$ determined by ${\sf exp}$, $c_{\sf mant}\in\mathbb{R}$ determined by ${\sf mant}$, whose precise definitions can be found in Sec. \ref{subsec:Quantization error bounding}.
%
A more principled approach to gradient quantization can be considered; see \cite{salehkalaibar2022lossy} for example. 
Here, we focus on fp conversion as it can be implemented with extreme computational efficiency.
%

Let us denote the fp quantization strategy as $Q_{\mathsf{fp}}(\cdot)$ in the following. 

\smallskip
\noindent
{\bf (ii) Huffman coding:} 
After fp conversion, the quantized gradient $\ghv^{(u)}_t$ is compressed using Huffman coding. 
%
As Huffman coding requires the distribution of data sources, we assume that the underlying distribution is $\gennorm$. 
Note that this assumption has been validated in Section~\ref{sec:Gradient modelling} for distributed DNN training without error correction. 
In Section~\ref{subsec:error_feedback}, we will verify this assumption again through simulations and obtain the corresponding parameters for distributed DNN with error feedback.
%
A different code is used in each DNN layer, but the same code is used across all users at a given layer.
Note that, as for the fp conversion, the Huffman coding is chosen for $\coiii$ as it can be implemented with minimum requirements for both computation and memory.
This is in contrast to other universal compression algorithms, such as LZ74 \cite{ziv1978compression}, which do not rely on any assumption on the source distribution.
A comparison in terms of gradient compression ratio between these two algorithms will be shown in Section~\ref{sec:simulation}.

In the following, we indicate the Huffman lossless compression as $h_{\mathsf{Hf}}(\cdot)$.

\smallskip
\noindent
{\bf (iii) Error correction:}
Motivated by the success of error feedback in accelerating the convergence of models using compressed gradients \cite{stich2018sparsified,stich2020error}, we adopt $f$ in Sec. \ref{sec:Rate-limited distributed training} that accumulates the quantization error from the previous epoch in the proposed $\coiii$. More precisely, for each $u\in[U]$ and a fixed $\gamma>0$, we let 
\ea{
    f(\mathbf{g}_1^{(u)},\ldots,\mathbf{g}_t^{(u)})=\mathbf{g}_t^{(u)}+\gamma \mathbf{m}_{t-1}^{(u)},\label{eq:f_error_fb}
}
where $\mathbf{m}_0^{(u)}=\zerov$ and
\ea{
    \mathbf{m}_t^{(u)}=\gamma\mathbf{m}_{t-1}^{(u)}+\mathbf{g}_t^{(u)}-\ghv^{(u)}_t,
}
with $\ghv^{(u)}_t = Q_{\mathsf{fp}}(f(\mathbf{g}_1^{(u)},\ldots,\mathbf{g}_t^{(u)}))$ being the quantization output. It is worth mentioning that \eqref{eq:f_error_fb} depends on $\mathbf{g}_1^{(u)},\ldots,\mathbf{g}_{t-1}^{(u)}$ only through $\mathbf{m}_{t-1}^{(u)}$; therefore, the algorithm can be easily implemented with limited memory resources. The parameter $\gamma$, which we refer to as \emph{memory decay coefficient}, is used to discount the error accumulation and our numerical experiments in Section~\ref{subsec:error_feedback} will show that a judicious choice of $\gamma$ is crucial in tuning performance. Also, to the best of our knowledge, this idea has not yet been adapted to schemes that leverage the statistical assumption of gradients; therefore, its effectiveness in such a scheme has not yet been confirmed.

We summarize the proposed $\coiii$ in Algorithm~\ref{alg:cap}, where Line 2 is to initialize $\mathbf{m}_0^{(u)}$; Line 8 first adds a version of quantization error from the previous epoch to the local gradient and performs quantization; Line 9 further compresses the output via a Huffman code; and Line 10 computes the quantization error of the current epoch.
A summary of the parameters in the proposed approach is provided in Table. \ref{tab:algo parameters}.

\begin{algorithm}
\caption{Proposed Algorithm: $\coiii$. Notations are provided both in Tab. \ref{tab:algo parameters} and Sec. \ref{subsec:CO}. }
\label{alg:cap}
\begin{algorithmic}[1]
\Require Local datasets $\{\Dcal_u \}_{u \in [U]}$, loss function $\Lcal(\cdot)$, initial model estimate $\whv_0$ 
\Require learning parameter $\eta$, memory decay parameter $\gamma$
\For{$u \in [U]$}
\State user $u$ sets memory to zero $\mv_0^{(u)}=\zerov$
\EndFor
\For{$t \in [T]$}
\State PS sends $\whv_t$ to all remote users
\For{$u \in [U]$}
\State user $u$ evaluates the local stochastic gradient $\gv^{(u)}_t$
\State user $u$  fp-converts $\gv^{(u)}$: $\ghv^{(u)}_t=Q_{\mathsf{fp}}(\gv^{(u)}_t+ \gamma \mv_{t-1}^{(u)} )$
\State user $u$ compresses  $\ghv^{(u)}_t$:  $\bv_t^{(u)}=h_{\mathsf{Hf}}(\ghv^{(u)}_t)$
\State user $u$ updates $\mv_t^{(u)}=\gamma \mv_{t-1}^{(u)} + \gv^{(u)}_t-\ghv^{(u)}_t $
\State user $u$ sends $\bv_t^{(u)}$ to the PS
\EndFor
\State PS decompresses all the users gradients as $\{\ghv^{(u)}_t\}_{u \in [U]}$
\State PS updates the model as $\whv_{t+1}=\whv_t-\f {\eta} U \sum_{u \in [U]} \ghv^{(u)}_t$
\EndFor
\State \Return $\whv_{T+1}$ an estimate of the optimal model $\wv^*$
\end{algorithmic}
\end{algorithm}

\begin{table}
	\footnotesize
	\centering
	\vspace{0.04in}\caption{Summary of key parameters in alphabetical order.}
	\label{tab:algo parameters}
	\begin{tabular}{|c|c|}
		\hline
  		Exponent Bias & $b$ \\ \hline
		Exponent Size & $\sf exp$ \\ \hline
	    fp Quantization & $Q_{\mathsf{fp}}(\cdot)$ \\ \hline
	    Stochastic Gradient & $\gv$ \\ \hline
	    Huffman Lossless Compression & $h_{\mathsf{Hf}}(\cdot)$ \\ \hline
	    Learning Rate & $\eta$ \\ \hline
	    Mantissa Size & $\sf mant$ \\ \hline
	    Memory & $\mv$ \\ \hline
	    Memory Decay Coefficient & $\gamma$ \\ \hline
	    Sign & $\sf sgn$ \\ \hline
        Total Time & $T$ \\ \hline
	    Total Users & $U$ \\ \hline
	\end{tabular}
	\vspace{-0.5cm}
\end{table}

\begin{remark}[Inconsequential number of users]
Note that in the approach of Algorithm \ref{alg:cap}, the number of remote users $U$  does not influence the accuracy/payload tradeoff in \eqref{eq:Lsf}. 
This is because the  algorithm parameters are not chosen as a function of $U$. In actuality, one would indeed design these hyper-parameters as a function of the number of remote users. 
 \end{remark}
Given the remark above, 
{\it we suppress the superscript ${(u)}$ from this point onward,} due to the fact that the user index is inconsequential.

\subsection{$\coiii$ for DNN}
\label{subsec:CO_DNN}
We now specialize the proposed $\coiii$ to DNN models. For our numerical evaluations, we consider the CIFAR-10 dataset classification task using the three architectures and training configurations specified earlier in Sec. \ref{sec:DNN training}.\footnote{We note that our following discussions are valid for all these three networks. However, to avoid repeating ourselves too many times, only a subset of the results is presented.}
%
%
%



Let us begin by revisiting the fp exponent bias in \eqref{eq:exp bias}. Two $Q_{\mathsf{fp}}(\cdot)$ are considered, namely $\sf[sgn, exp, mant]=[1,2,1]$ and $\sf[sgn, exp, mant]=[1,5,2]$, which we refer to as fp4 and fp8, respectively, in the sequel. For each $Q_{\mathsf{fp}}(\cdot)$, we note that solving \eqref{eq:exp bias} explicitly is not easy at all due to the complicated nature of $\gennorm$ distribution (which is evident by the fact that the rate-distortion problem remains unsolved for most $\beta$ \cite{Fraysse08}) and that of the fp quantizers. 

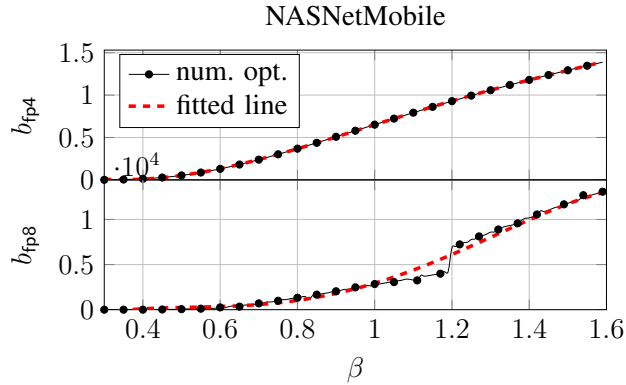
\begin{figure}
    \centering


\begin{tikzpicture}[thick, scale=0.9]
    \begin{groupplot}[
        group style={
            group name=my plots,
            group size=1 by 2,
            xlabels at=edge bottom,
            xticklabels at=edge bottom,
            vertical sep=0pt
        },
        height=3.5cm,
        width=9cm,
        ymin=0,
        xmin=0.3,
        xmax=1.6,
        xlabel={$\beta$},
        legend pos=north west,
        grid=both
    ]
    \nextgroupplot[title=NASNetMobile,
    ylabel={$b_{\sf fp4}$}]
        \coordinate (top) at (axis cs:1,\pgfkeysvalueof{/pgfplots/ymax});
        \addplot [only marks,color=black,mark=*,mark options={solid},mark repeat={5},mark size=1.5pt, smooth]
            table[x index=0, y index=1]{./Data/gennorm_beta_scalar_function.txt};
            \addlegendentry{num. opt.}
        \addplot [dashed, draw=red, line width=1.5pt, smooth]
            table[x index=0, y index=2]{./Data/gennorm_beta_scalar_function.txt};
            \addlegendentry{fitted line}
    \nextgroupplot[ylabel style={align=center},
    ylabel={$b_{\sf fp8}$}]
        \addplot [only marks,color=black,mark=*,mark options={solid},mark repeat={5},mark size=1.5pt, smooth]
            table[x index=0, y index=1]{./Data/fp8_best_scalar_function.txt};
        \addplot [dashed, draw=red, line width=1.5pt, smooth]
            table[x index=0, y index=2]{./Data/fp8_best_scalar_function.txt};
            
    \coordinate (bot) at (axis cs:1,\pgfkeysvalueof{/pgfplots/ymin});
    \end{groupplot}
    
    %
\end{tikzpicture}
    \vspace{-0.5cm}
    \caption{The relation between $\beta$ and $b$ which minimizes the $L_2$ loss for fp4 and fp8 quantization.}
    \label{fig:beta_scalar_function}
	\vspace{-0.5cm}
\end{figure}

Here, we instead solve \eqref{eq:exp bias} via Monte-Carlo simulations for some isolated instances of $\beta$ and then look for a polynomial to fit these simulated results. As a result, we are able to argue that, when the $\gennorm$ assumption holds, then  
\ea{
b_t \approxeq  \lb 0.46 -2.85\beta + 5.37\beta^2  -2.85 \beta^3 + 0.52 \beta^4 \rb /\sigma ,
\label{eq:approximate}
}
where $\approxeq$ is used to represent that the left-hand side and the right-hand side are approximately equal to each other, $\beta$ is the $\gennorm$ shape parameter corresponding to the given DNN layer, and $\sigma^2$ is the variance.
In other words, $b_t$ can be well-approximated with a polynomial that depends only on the shape parameter, once normalized by the variance.
In the top of Fig. \ref{fig:beta_scalar_function}, we plot the numerically optimized $b_t$ for fp4 as a function of $\beta$ together with the approximation in \eqref{eq:approximate} for the case in which the variance is unitary. 

Similar tests are done for fp8 for which we obtain
\begin{align}
    b_t &\approxeq  \left( - 5793 +35605.5\beta -76511.8\beta^2\right. \nonumber\\  
    &\left.\hspace{1in}+68153 \beta^3 -18520.3 \beta^4 \right)/\sigma.
    \label{eq:approximate_fp8}
\end{align}
Moreover, the numerically optimized $b_t$ for fp8 and the approximation in \eqref{eq:approximate_fp8} are shown in the bottom of Fig.~\ref{fig:beta_scalar_function}. In all of the following simulations, we will use \eqref{eq:approximate} and \eqref{eq:approximate_fp8} to select $b_t$ for fp4 and fp8, respectively.

%
%

%
%

\subsection{Validation of $\gennorm$ assumption with error feedback}
\label{subsec:error_feedback}
\begin{figure*}[t]
    \centering
	\begin{tikzpicture}[thick, scale=0.8]
    \definecolor{mycolor1}{rgb}{0.00000,0.44706,0.74118}%
    \definecolor{mycolor2}{rgb}{0.63529,0.07843,0.18431}%
    \definecolor{mycolor3}{rgb}{0.00000,0.49804,0.00000}%
    \definecolor{mycolor4}{rgb}{0.60000,0.19608,0.80000}%
    \begin{groupplot}[
        group style={
            group name=my plots,
            group size=2 by 3,
            xlabels at=edge bottom,
            xticklabels at=edge bottom,
            vertical sep=0pt
        },
        height=3.5cm, 
        width=9cm, 
        xmin=0,
        xlabel=Epoch,
        grid=both
    ]
    \nextgroupplot[title=$W_2$ distance,xmax=150,
    ylabel=Upper]
        \coordinate (top) at (axis cs:1,\pgfkeysvalueof{/pgfplots/ymax});
        \addplot [draw=mycolor1, line width=1.5pt, mark=diamond, mark options={solid, mycolor1}, mark repeat={2},each nth point=5,smooth]
            table[x index=0, y index=1]{./Data/nasnet_0.7_upper_w2.txt};
            \label{plots:plot1_nasnet_w2}
        \addplot [draw=mycolor2, line width=1.5pt, mark=diamond, mark options={solid, mycolor2}, mark repeat={2},each nth point=5, smooth]
            table[x index=0, y index=2]{./Data/nasnet_0.7_upper_w2.txt};
            \label{plots:plot2_nasnet_w2}
        \addplot [draw=mycolor3, line width=1.5pt, mark=square, mark options={solid, mycolor3}, mark repeat={2},each nth point=5, smooth]
            table[x index=0, y index=3]{./Data/nasnet_0.7_upper_w2.txt};
            \label{plots:plot3_nasnet_w2}
        \addplot [draw=mycolor4, line width=1.5pt, mark=square, mark options={solid, mycolor4}, mark repeat={2},each nth point=5, smooth]
            table[x index=0, y index=4]{./Data/nasnet_0.7_upper_w2.txt};
            \label{plots:plot4_nasnet_w2}

    \nextgroupplot[title=$W_2$ distance ,xmax=50]
        \coordinate (top) at (axis cs:1,\pgfkeysvalueof{/pgfplots/ymax});
        \addplot [draw=mycolor1, line width=1.5pt, mark=diamond, mark options={solid, mycolor1}, mark repeat={2},each nth point=2,smooth]
            table[x index=0, y index=1]{./Data/teacher_student/CO3/fp4/gamma0.7/NASNetMobile_upper_w2.txt};
            \label{plots:plot1_nasnet_w2}
        \addplot [draw=mycolor2, line width=1.5pt, mark=diamond, mark options={solid, mycolor2}, mark repeat={2},each nth point=2, smooth]
            table[x index=0, y index=2]{./Data/teacher_student/CO3/fp4/gamma0.7/NASNetMobile_upper_w2.txt};
            \label{plots:plot2_nasnet_w2}
        \addplot [draw=mycolor3, line width=1.5pt, mark=square, mark options={solid, mycolor3}, mark repeat={2},each nth point=2, smooth]
            table[x index=0, y index=3]{./Data/teacher_student/CO3/fp4/gamma0.7/NASNetMobile_upper_w2.txt};
            \label{plots:plot3_nasnet_w2}
        \addplot [draw=mycolor4, line width=1.5pt, mark=square, mark options={solid, mycolor4}, mark repeat={2},each nth point=2, smooth]
            table[x index=0, y index=4]{./Data/teacher_student/CO3/fp4/gamma0.9/NASNetMobile_upper_w2.txt};
            \label{plots:plot4_nasnet_w2}
            
    \nextgroupplot[ylabel style={align=center},xmax=150,
    ylabel=Middle]
        \addplot [draw=mycolor1, line width=1.5pt, mark=diamond, mark options={solid, mycolor1}, mark repeat={2},each nth point=5, smooth]
            table[x index=0, y index=1]{./Data/nasnet_0.7_middle_w2.txt};
        \addplot [draw=mycolor2, line width=1.5pt, mark=diamond, mark options={solid, mycolor2}, mark repeat={2},each nth point=5, smooth]
            table[x index=0, y index=2]{./Data/nasnet_0.7_middle_w2.txt};
        \addplot [draw=mycolor3, line width=1.5pt, mark=square, mark options={solid, mycolor3}, mark repeat={2},each nth point=5, smooth]
            table[x index=0, y index=3]{./Data/nasnet_0.7_middle_w2.txt};
        \addplot [draw=mycolor4, line width=1.5pt, mark=square, mark
        options={solid, mycolor4},each nth point=5, mark repeat={10}, smooth]
            table[x index=0, y index=4]{./Data/nasnet_0.7_middle_w2.txt};

    \nextgroupplot[ylabel style={align=center},xmax=50]
        \addplot [draw=mycolor1, line width=1.5pt, mark=diamond, mark options={solid, mycolor1}, mark repeat={2},each nth point=2, smooth]
            table[x index=0, y index=1]{./Data/teacher_student/CO3/fp4/gamma0.7/NASNetMobile_middle_w2.txt};
        \addplot [draw=mycolor2, line width=1.5pt, mark=diamond, mark options={solid, mycolor2}, mark repeat={2},each nth point=2, smooth]
            table[x index=0, y index=2]{./Data/teacher_student/CO3/fp4/gamma0.7/NASNetMobile_middle_w2.txt};
        \addplot [draw=mycolor3, line width=1.5pt, mark=square, mark options={solid, mycolor3}, mark repeat={2},each nth point=2, smooth]
            table[x index=0, y index=3]{./Data/teacher_student/CO3/fp4/gamma0.7/NASNetMobile_middle_w2.txt};
        \addplot [draw=mycolor4, line width=1.5pt, mark=square, mark options={solid, mycolor4}, mark repeat={2},each nth point=2, smooth]
            table[x index=0, y index=4]{./Data/teacher_student/CO3/fp4/gamma0.7/NASNetMobile_middle_w2.txt};
    
    \nextgroupplot[ylabel style={align=center},xmax=150,
    ylabel=Lower]
        \addplot [draw=mycolor1, line width=1.5pt, mark=diamond, mark options={solid, mycolor1}, mark repeat={2},each nth point=5, smooth]
            table[x index=0, y index=1]{./Data/nasnet_0.7_lower_w2.txt};
        \addplot [draw=mycolor2, line width=1.5pt, mark=diamond, mark options={solid, mycolor2}, mark repeat={2},each nth point=5, smooth]
            table[x index=0, y index=2]{./Data/nasnet_0.7_lower_w2.txt};
        \addplot [draw=mycolor3, line width=1.5pt, mark=square, mark options={solid, mycolor3}, mark repeat={2},each nth point=5, smooth]
            table[x index=0, y index=3]{./Data/nasnet_0.7_lower_w2.txt};
        \addplot [draw=mycolor4, line width=1.5pt, mark=square, mark options={solid, mycolor4}, mark repeat={2},each nth point=5, smooth]
            table[x index=0, y index=4]{./Data/nasnet_0.7_lower_w2.txt};

    \nextgroupplot[ylabel style={align=center},xmax=50]
        \addplot [draw=mycolor1, line width=1.5pt, mark=diamond, mark options={solid, mycolor1}, mark repeat={2},each nth point=2, smooth]
            table[x index=0, y index=1]{./Data/teacher_student/CO3/fp4/gamma0.7/NASNetMobile_lower_w2.txt};
        \addplot [draw=mycolor2, line width=1.5pt, mark=diamond, mark options={solid, mycolor2}, mark repeat={2},each nth point=2, smooth]
            table[x index=0, y index=2]{./Data/teacher_student/CO3/fp4/gamma0.7/NASNetMobile_lower_w2.txt};
        \addplot [draw=mycolor3, line width=1.5pt, mark=square, mark options={solid, mycolor3}, mark repeat={2},each nth point=2, smooth]
            table[x index=0, y index=3]{./Data/teacher_student/CO3/fp4/gamma0.7/NASNetMobile_lower_w2.txt};
        \addplot [draw=mycolor4, line width=1.5pt, mark=square, mark options={solid, mycolor4}, mark repeat={2},each nth point=2, smooth]
            table[x index=0, y index=4]{./Data/teacher_student/CO3/fp4/gamma0.7/NASNetMobile_lower_w2.txt};
            
    \coordinate (bot) at (axis cs:1,\pgfkeysvalueof{/pgfplots/ymin});
    \end{groupplot}

    \path (top|-current bounding box.north)--
          coordinate(legendpos)
          (bot|-current bounding box.north);
    \matrix[
        matrix of nodes,
        anchor=south,
        draw,
        inner sep=0.2em,
        draw
      ]at([yshift=1ex,xshift=-5ex]legendpos)
      {
        \ref{plots:plot1_nasnet_w2}& normal &[3pt]
        \ref{plots:plot2_nasnet_w2}& Laplace &[3pt]
        \ref{plots:plot3_nasnet_w2}& dWeibull &[3pt]
        \ref{plots:plot4_nasnet_w2}& $\gennorm$ &[3pt]\\};
\end{tikzpicture}
    \vspace{-0.5cm}
    \caption{$W_2$ distance between the empirical  CDF and best-fit CDF for the term $\gv_t+0.7 \mv_{t-1}^{(u)}$ with fp4 training for each layer of NASNetMobile.
    CIFAR10 classification task (left) and teacher-student setting (right).}
    \label{fig:nasnetmobile_gamma0.7_w2}
    \vspace{-0.5cm}
\end{figure*}
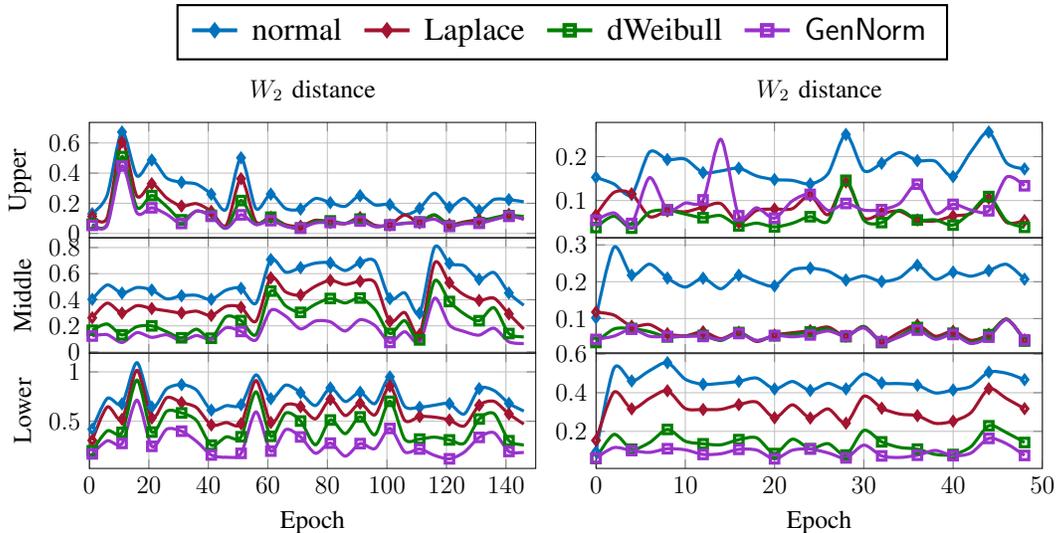

Next, we wish to validate the $\gennorm$ assumption of Section~\ref{sec:Gradient modelling} even when error correction is employed, that is when $\gv_t+\gamma \mv_{t-1}^{(u)}$ is considered.  
In Figs. \ref{fig:nasnetmobile_gamma0.9_w2} and \ref{fig:nasnetmobile_gamma0.7_w2}, we consider fp4 quantization and $\gamma=0.9$ and $0.7$, respectively, for error correction in $\coiii$ and plot $W_2$ the Wasserstein distance of order 2 between the empirical CDF and the best-fit CDFs in the four families: (i) normal, (ii) Laplace (proposed in \cite{berivan2021laplace} for modeling DNN gradients), (iii) double Weibull (proposed in \cite{fangcheng2020tinyscript} for modeling DNN gradients), and (iv) $\gennorm$. 
%
%
\begin{figure*}[t]
    \centering
	\begin{tikzpicture}[thick, scale=0.8]
    \definecolor{mycolor1}{rgb}{0.00000,0.44706,0.74118}%
    \definecolor{mycolor2}{rgb}{0.63529,0.07843,0.18431}%
    \definecolor{mycolor3}{rgb}{0.00000,0.49804,0.00000}%
    \definecolor{mycolor4}{rgb}{0.60000,0.19608,0.80000}%
    \begin{groupplot}[
        group style={
            group name=my plots,
            group size=2 by 3,
            xlabels at=edge bottom,
            xticklabels at=edge bottom,
            vertical sep=0pt
        },
        height=3.5cm, 
        width=9cm, 
        xmin=0,
        xlabel=Epoch,
        grid=both
    ]
    \nextgroupplot[title=$W_2$ distance,xmax=150,
    ylabel=Upper]
        \coordinate (top) at (axis cs:1,\pgfkeysvalueof{/pgfplots/ymax});
        \addplot [draw=mycolor1, line width=1.5pt, mark=diamond, mark options={solid, mycolor1}, mark repeat={2},each nth point=5,smooth]
            table[x index=0, y index=1]{./Data/cifar10/fp4/gamma0.9/NASNetMobile_upper_w2.txt};
            \label{plots:plot1_nasnet_w2}
        \addplot [draw=mycolor2, line width=1.5pt, mark=diamond, mark options={solid, mycolor2}, mark repeat={2},each nth point=5, smooth]
            table[x index=0, y index=2]{./Data/cifar10/fp4/gamma0.9/NASNetMobile_upper_w2.txt};
            \label{plots:plot2_nasnet_w2}
        \addplot [draw=mycolor3, line width=1.5pt, mark=square, mark options={solid, mycolor3}, mark repeat={2},each nth point=5, smooth]
            table[x index=0, y index=3]{./Data/cifar10/fp4/gamma0.9/NASNetMobile_upper_w2.txt};
            \label{plots:plot3_nasnet_w2}
        \addplot [draw=mycolor4, line width=1.5pt, mark=square, mark options={solid, mycolor4}, mark repeat={2},each nth point=5, smooth]
            table[x index=0, y index=4]{./Data/cifar10/fp4/gamma0.9/NASNetMobile_upper_w2.txt};
            \label{plots:plot4_nasnet_w2}

    \nextgroupplot[title=$W_2$ distance ,xmax=50]
        \coordinate (top) at (axis cs:1,\pgfkeysvalueof{/pgfplots/ymax});
        \addplot [draw=mycolor1, line width=1.5pt, mark=diamond, mark options={solid, mycolor1}, mark repeat={2},each nth point=2,smooth]
            table[x index=0, y index=1]{./Data/teacher_student/CO3/fp4/gamma0.9/NASNetMobile_upper_w2.txt};
            \label{plots:plot1_nasnet_w2}
        \addplot [draw=mycolor2, line width=1.5pt, mark=diamond, mark options={solid, mycolor2}, mark repeat={2},each nth point=2, smooth]
            table[x index=0, y index=2]{./Data/teacher_student/CO3/fp4/gamma0.9/NASNetMobile_upper_w2.txt};
            \label{plots:plot2_nasnet_w2}
        \addplot [draw=mycolor3, line width=1.5pt, mark=square, mark options={solid, mycolor3}, mark repeat={2},each nth point=2, smooth]
            table[x index=0, y index=3]{./Data/teacher_student/CO3/fp4/gamma0.9/NASNetMobile_upper_w2.txt};
            \label{plots:plot3_nasnet_w2}
        \addplot [draw=mycolor4, line width=1.5pt, mark=square, mark options={solid, mycolor4}, mark repeat={2},each nth point=2, smooth]
            table[x index=0, y index=4]{./Data/teacher_student/CO3/fp4/gamma0.9/NASNetMobile_upper_w2.txt};
            \label{plots:plot4_nasnet_w2}
            
    \nextgroupplot[ylabel style={align=center},xmax=150,
    ylabel=Middle]
        \addplot [draw=mycolor1, line width=1.5pt, mark=diamond, mark options={solid, mycolor1}, mark repeat={2},each nth point=5, smooth]
            table[x index=0, y index=1]{./Data/cifar10/fp4/gamma0.9/NASNetMobile_middle_w2.txt};
        \addplot [draw=mycolor2, line width=1.5pt, mark=diamond, mark options={solid, mycolor2}, mark repeat={2},each nth point=5, smooth]
            table[x index=0, y index=2]{./Data/cifar10/fp4/gamma0.9/NASNetMobile_middle_w2.txt};
        \addplot [draw=mycolor3, line width=1.5pt, mark=square, mark options={solid, mycolor3}, mark repeat={2},each nth point=5, smooth]
            table[x index=0, y index=3]{./Data/cifar10/fp4/gamma0.9/NASNetMobile_middle_w2.txt};
        \addplot [draw=mycolor4, line width=1.5pt, mark=square, mark
        options={solid, mycolor4},each nth point=5, mark repeat={10}, smooth]
            table[x index=0, y index=4]{./Data/cifar10/fp4/gamma0.9/NASNetMobile_middle_w2.txt};

    \nextgroupplot[ylabel style={align=center},xmax=50]
        \addplot [draw=mycolor1, line width=1.5pt, mark=diamond, mark options={solid, mycolor1}, mark repeat={2},each nth point=2, smooth]
            table[x index=0, y index=1]{./Data/teacher_student/CO3/fp4/gamma0.9/NASNetMobile_middle_w2.txt};
        \addplot [draw=mycolor2, line width=1.5pt, mark=diamond, mark options={solid, mycolor2}, mark repeat={2},each nth point=2, smooth]
            table[x index=0, y index=2]{./Data/teacher_student/CO3/fp4/gamma0.9/NASNetMobile_middle_w2.txt};
        \addplot [draw=mycolor3, line width=1.5pt, mark=square, mark options={solid, mycolor3}, mark repeat={2},each nth point=2, smooth]
            table[x index=0, y index=3]{./Data/teacher_student/CO3/fp4/gamma0.9/NASNetMobile_middle_w2.txt};
        \addplot [draw=mycolor4, line width=1.5pt, mark=square, mark options={solid, mycolor4}, mark repeat={2},each nth point=2, smooth]
            table[x index=0, y index=4]{./Data/teacher_student/CO3/fp4/gamma0.9/NASNetMobile_middle_w2.txt};
    
    \nextgroupplot[ylabel style={align=center},xmax=150,
    ylabel=Lower]
        \addplot [draw=mycolor1, line width=1.5pt, mark=diamond, mark options={solid, mycolor1}, mark repeat={2},each nth point=5, smooth]
            table[x index=0, y index=1]{./Data/cifar10/fp4/gamma0.9/NASNetMobile_lower_w2.txt};
        \addplot [draw=mycolor2, line width=1.5pt, mark=diamond, mark options={solid, mycolor2}, mark repeat={2},each nth point=5, smooth]
            table[x index=0, y index=2]{./Data/cifar10/fp4/gamma0.9/NASNetMobile_lower_w2.txt};
        \addplot [draw=mycolor3, line width=1.5pt, mark=square, mark options={solid, mycolor3}, mark repeat={2},each nth point=5, smooth]
            table[x index=0, y index=3]{./Data/cifar10/fp4/gamma0.9/NASNetMobile_lower_w2.txt};
        \addplot [draw=mycolor4, line width=1.5pt, mark=square, mark options={solid, mycolor4}, mark repeat={2},each nth point=5, smooth]
            table[x index=0, y index=4]{./Data/cifar10/fp4/gamma0.9/NASNetMobile_lower_w2.txt};

    \nextgroupplot[ylabel style={align=center},xmax=50]
        \addplot [draw=mycolor1, line width=1.5pt, mark=diamond, mark options={solid, mycolor1}, mark repeat={2},each nth point=2, smooth]
            table[x index=0, y index=1]{./Data/teacher_student/CO3/fp4/gamma0.9/NASNetMobile_lower_w2.txt};
        \addplot [draw=mycolor2, line width=1.5pt, mark=diamond, mark options={solid, mycolor2}, mark repeat={2},each nth point=2, smooth]
            table[x index=0, y index=2]{./Data/teacher_student/CO3/fp4/gamma0.9/NASNetMobile_lower_w2.txt};
        \addplot [draw=mycolor3, line width=1.5pt, mark=square, mark options={solid, mycolor3}, mark repeat={2},each nth point=2, smooth]
            table[x index=0, y index=3]{./Data/teacher_student/CO3/fp4/gamma0.9/NASNetMobile_lower_w2.txt};
        \addplot [draw=mycolor4, line width=1.5pt, mark=square, mark options={solid, mycolor4}, mark repeat={2},each nth point=2, smooth]
            table[x index=0, y index=4]{./Data/teacher_student/CO3/fp4/gamma0.9/NASNetMobile_lower_w2.txt};
            
    \coordinate (bot) at (axis cs:1,\pgfkeysvalueof{/pgfplots/ymin});
    \end{groupplot}

    \path (top|-current bounding box.north)--
          coordinate(legendpos)
          (bot|-current bounding box.north);
    \matrix[
        matrix of nodes,
        anchor=south,
        draw,
        inner sep=0.2em,
        draw
      ]at([yshift=1ex,xshift=-5ex]legendpos)
      {
        \ref{plots:plot1_nasnet_w2}& normal &[3pt]
        \ref{plots:plot2_nasnet_w2}& Laplace &[3pt]
        \ref{plots:plot3_nasnet_w2}& dWeibull &[3pt]
        \ref{plots:plot4_nasnet_w2}& $\gennorm$ &[3pt]\\};
\end{tikzpicture}
    \vspace{-0.5cm}
    \caption{$W_2$ distance between the empirical  CDF and best-fit CDF for the term $\gv_t+0.9 \mv_{t-1}^{(u)}$ with fp4 training for each layer of NASNetMobile.
    CIFAR10 classification task (left) and teacher-student setting (right).}
    \label{fig:nasnetmobile_gamma0.9_w2}
    \vspace{-0.5cm}
\end{figure*}
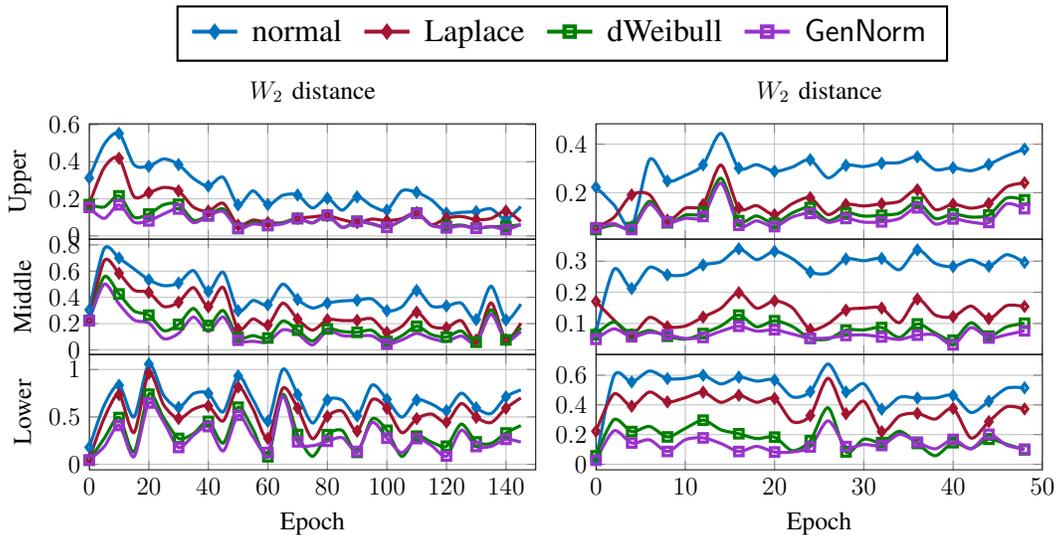

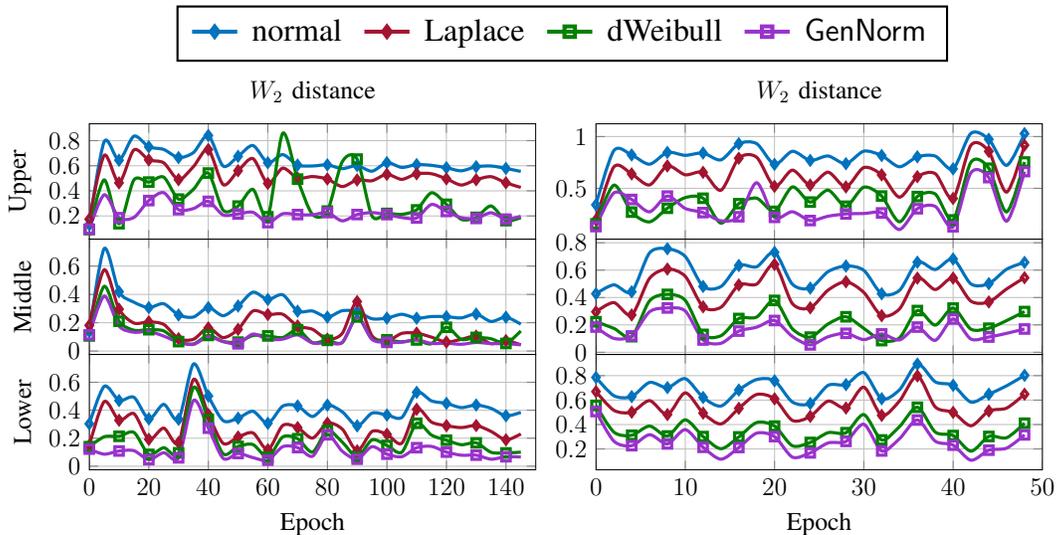
\begin{figure*}[t]
    \centering
	\begin{tikzpicture}[thick, scale=0.8]
    \definecolor{mycolor1}{rgb}{0.00000,0.44706,0.74118}%
    \definecolor{mycolor2}{rgb}{0.63529,0.07843,0.18431}%
    \definecolor{mycolor3}{rgb}{0.00000,0.49804,0.00000}%
    \definecolor{mycolor4}{rgb}{0.60000,0.19608,0.80000}%
    \begin{groupplot}[
        group style={
            group name=my plots,
            group size=2 by 3,
            xlabels at=edge bottom,
            xticklabels at=edge bottom,
            vertical sep=0pt
        },
        height=3.5cm, 
        width=9cm, 
        xmin=0,
        xlabel=Epoch,
        grid=both
    ]
    \nextgroupplot[title=$W_2$ distance,xmax=150,
    ylabel=Upper]
        \coordinate (top) at (axis cs:1,\pgfkeysvalueof{/pgfplots/ymax});
        \addplot [draw=mycolor1, line width=1.5pt, mark=diamond, mark options={solid, mycolor1}, mark repeat={2},each nth point=5,smooth]
            table[x index=0, y index=1]{./Data/ResNet50V2_8bit_upper_w2.txt};
            \label{plots:plot1_nasnet_w2}
        \addplot [draw=mycolor2, line width=1.5pt, mark=diamond, mark options={solid, mycolor2}, mark repeat={2},each nth point=5, smooth]
            table[x index=0, y index=2]{./Data/ResNet50V2_8bit_upper_w2.txt};
            \label{plots:plot2_nasnet_w2}
        \addplot [draw=mycolor3, line width=1.5pt, mark=square, mark options={solid, mycolor3}, mark repeat={2},each nth point=5, smooth]
            table[x index=0, y index=3]{./Data/ResNet50V2_8bit_upper_w2.txt};
            \label{plots:plot3_nasnet_w2}
        \addplot [draw=mycolor4, line width=1.5pt, mark=square, mark options={solid, mycolor4}, mark repeat={2},each nth point=5, smooth]
            table[x index=0, y index=4]{./Data/ResNet50V2_8bit_upper_w2.txt};
            \label{plots:plot4_nasnet_w2}

    \nextgroupplot[title=$W_2$ distance ,xmax=50]
        \coordinate (top) at (axis cs:1,\pgfkeysvalueof{/pgfplots/ymax});
        \addplot [draw=mycolor1, line width=1.5pt, mark=diamond, mark options={solid, mycolor1}, mark repeat={2},each nth point=2,smooth]
            table[x index=0, y index=1]{./Data/teacher_student/CO3/fp8/gamma0.7/ResNet50V2_upper_w2.txt};
            \label{plots:plot1_nasnet_w2}
        \addplot [draw=mycolor2, line width=1.5pt, mark=diamond, mark options={solid, mycolor2}, mark repeat={2},each nth point=2, smooth]
            table[x index=0, y index=2]{./Data/teacher_student/CO3/fp8/gamma0.7/ResNet50V2_upper_w2.txt};
            \label{plots:plot2_nasnet_w2}
        \addplot [draw=mycolor3, line width=1.5pt, mark=square, mark options={solid, mycolor3}, mark repeat={2},each nth point=2, smooth]
            table[x index=0, y index=3]{./Data/teacher_student/CO3/fp8/gamma0.7/ResNet50V2_upper_w2.txt};
            \label{plots:plot3_nasnet_w2}
        \addplot [draw=mycolor4, line width=1.5pt, mark=square, mark options={solid, mycolor4}, mark repeat={2},each nth point=2, smooth]
            table[x index=0, y index=4]{./Data/teacher_student/CO3/fp8/gamma0.7/ResNet50V2_upper_w2.txt};
            \label{plots:plot4_nasnet_w2}
            
    \nextgroupplot[ylabel style={align=center},xmax=150,
    ylabel=Middle]
        \addplot [draw=mycolor1, line width=1.5pt, mark=diamond, mark options={solid, mycolor1}, mark repeat={2},each nth point=5, smooth]
            table[x index=0, y index=1]{./Data/ResNet50V2_8bit_middle_w2.txt};
        \addplot [draw=mycolor2, line width=1.5pt, mark=diamond, mark options={solid, mycolor2}, mark repeat={2},each nth point=5, smooth]
            table[x index=0, y index=2]{./Data/ResNet50V2_8bit_middle_w2.txt};
        \addplot [draw=mycolor3, line width=1.5pt, mark=square, mark options={solid, mycolor3}, mark repeat={2},each nth point=5, smooth]
            table[x index=0, y index=3]{./Data/ResNet50V2_8bit_middle_w2.txt};
        \addplot [draw=mycolor4, line width=1.5pt, mark=square, mark
        options={solid, mycolor4},each nth point=5, mark repeat={10}, smooth]
            table[x index=0, y index=4]{./Data/ResNet50V2_8bit_middle_w2.txt};

    \nextgroupplot[ylabel style={align=center},xmax=50]
        \addplot [draw=mycolor1, line width=1.5pt, mark=diamond, mark options={solid, mycolor1}, mark repeat={2},each nth point=2, smooth]
            table[x index=0, y index=1]{./Data/teacher_student/CO3/fp8/gamma0.7/ResNet50V2_middle_w2.txt};
        \addplot [draw=mycolor2, line width=1.5pt, mark=diamond, mark options={solid, mycolor2}, mark repeat={2},each nth point=2, smooth]
            table[x index=0, y index=2]{./Data/teacher_student/CO3/fp8/gamma0.7/ResNet50V2_middle_w2.txt};
        \addplot [draw=mycolor3, line width=1.5pt, mark=square, mark options={solid, mycolor3}, mark repeat={2},each nth point=2, smooth]
            table[x index=0, y index=3]{./Data/teacher_student/CO3/fp8/gamma0.7/ResNet50V2_middle_w2.txt};
        \addplot [draw=mycolor4, line width=1.5pt, mark=square, mark options={solid, mycolor4}, mark repeat={2},each nth point=2, smooth]
            table[x index=0, y index=4]{./Data/teacher_student/CO3/fp8/gamma0.7/ResNet50V2_middle_w2.txt};
    
    \nextgroupplot[ylabel style={align=center},xmax=150,
    ylabel=Lower]
        \addplot [draw=mycolor1, line width=1.5pt, mark=diamond, mark options={solid, mycolor1}, mark repeat={2},each nth point=5, smooth]
            table[x index=0, y index=1]{./Data/ResNet50V2_8bit_lower_w2.txt};
        \addplot [draw=mycolor2, line width=1.5pt, mark=diamond, mark options={solid, mycolor2}, mark repeat={2},each nth point=5, smooth]
            table[x index=0, y index=2]{./Data/ResNet50V2_8bit_lower_w2.txt};
        \addplot [draw=mycolor3, line width=1.5pt, mark=square, mark options={solid, mycolor3}, mark repeat={2},each nth point=5, smooth]
            table[x index=0, y index=3]{./Data/ResNet50V2_8bit_lower_w2.txt};
        \addplot [draw=mycolor4, line width=1.5pt, mark=square, mark options={solid, mycolor4}, mark repeat={2},each nth point=5, smooth]
            table[x index=0, y index=4]{./Data/ResNet50V2_8bit_lower_w2.txt};

    \nextgroupplot[ylabel style={align=center},xmax=50]
        \addplot [draw=mycolor1, line width=1.5pt, mark=diamond, mark options={solid, mycolor1}, mark repeat={2},each nth point=2, smooth]
            table[x index=0, y index=1]{./Data/teacher_student/CO3/fp8/gamma0.7/ResNet50V2_lower_w2.txt};
        \addplot [draw=mycolor2, line width=1.5pt, mark=diamond, mark options={solid, mycolor2}, mark repeat={2},each nth point=2, smooth]
            table[x index=0, y index=2]{./Data/teacher_student/CO3/fp8/gamma0.7/ResNet50V2_lower_w2.txt};
        \addplot [draw=mycolor3, line width=1.5pt, mark=square, mark options={solid, mycolor3}, mark repeat={2},each nth point=2, smooth]
            table[x index=0, y index=3]{./Data/teacher_student/CO3/fp8/gamma0.7/ResNet50V2_lower_w2.txt};
        \addplot [draw=mycolor4, line width=1.5pt, mark=square, mark options={solid, mycolor4}, mark repeat={2},each nth point=2, smooth]
            table[x index=0, y index=4]{./Data/teacher_student/CO3/fp8/gamma0.7/ResNet50V2_lower_w2.txt};
            
    \coordinate (bot) at (axis cs:1,\pgfkeysvalueof{/pgfplots/ymin});
    \end{groupplot}

    \path (top|-current bounding box.north)--
          coordinate(legendpos)
          (bot|-current bounding box.north);
    \matrix[
        matrix of nodes,
        anchor=south,
        draw,
        inner sep=0.2em,
        draw
      ]at([yshift=1ex,xshift=-5ex]legendpos)
      {
        \ref{plots:plot1_nasnet_w2}& normal &[3pt]
        \ref{plots:plot2_nasnet_w2}& Laplace &[3pt]
        \ref{plots:plot3_nasnet_w2}& dWeibull &[3pt]
        \ref{plots:plot4_nasnet_w2}& $\gennorm$ &[3pt]\\};
\end{tikzpicture}
    \vspace{-0.5cm}
    \caption{$W_2$ distance between the empirical  CDF and best-fit CDF for the term $\gv_t+0.7 \mv_{t-1}^{(u)}$ with fp8 training for each layer of ResNet50V2.
    CIFAR10 classification task (left) and teacher-student setting (right).}
    \label{fig:resnet50v2_fp8_gamma0.7_w2}
    \vspace{-0.5cm}
\end{figure*}

From the left of Fig. \ref{fig:nasnetmobile_gamma0.9_w2} and \ref{fig:nasnetmobile_gamma0.7_w2}, we observe that the $\gennorm$ distribution indeed yields in the smallest $W_2$ distance among the four families for CIFAR-10 classification task. The same results are shown in the right column of these two figures, which again validate the $\gennorm$ assumption even in the teacher-student setting.


%
%
In Fig.~\ref{fig:resnet50v2_fp8_gamma0.7_w2}, we repeat the simulation for ResNet50V2 but replace fp4 with fp8. In this figure, we can one again observe that among the four families of distributions, $\gennorm$ best matches the empirical distribution.






\subsection{Error magnitude}
\label{subsec:error_mag}
\begin{figure*}[t]
    \centering
	\begin{tikzpicture}[thick, scale=0.8]
    \definecolor{mycolor1}{rgb}{0.00000,0.44706,0.74118}%
    \definecolor{mycolor2}{rgb}{0.63529,0.07843,0.18431}%
    \definecolor{mycolor3}{rgb}{0.00000,0.49804,0.00000}%
    \definecolor{mycolor4}{rgb}{0.60000,0.19608,0.80000}%
    \begin{groupplot}[
        group style={
            group name=my plots,
            group size=2 by 3,
            xlabels at=edge bottom,
            xticklabels at=edge bottom,
            vertical sep=0pt
        },
        height=3.5cm, 
        width=9cm, 
        xmin=0,
        xlabel=Epoch,
        grid=both
    ]
    \nextgroupplot[title=ResNet50V2,
    ylabel=Upper]
        \coordinate (top) at (axis cs:1,\pgfkeysvalueof{/pgfplots/ymax});
        \addplot [draw=mycolor1, line width=1.5pt, mark=diamond, mark options={solid, mycolor1}, mark repeat={10}, smooth]
            table[x index=0, y index=1]{./Data/resnet50v2_gamma0.7_grad_l1norm.txt};
            \label{plots:plot1}
        \addplot [draw=mycolor2, line width=1.5pt, mark=square, mark options={solid, mycolor2}, mark repeat={10}, smooth]
            table[x index=0, y index=1]{./Data/resnet50v2_gamma0.7_error_l1norm.txt};
            \label{plots:plot2}

    \nextgroupplot[title=ResNet50V2,
    ylabel=Upper]
        \coordinate (top) at (axis cs:1,\pgfkeysvalueof{/pgfplots/ymax});
        \addplot [draw=mycolor1, line width=1.5pt, mark=diamond, mark options={solid, mycolor1}, mark repeat={10}, smooth]
            table[x index=0, y index=1]{./Data/resnet50v2_gamma0.9_grad_l1norm.txt};
            \label{plots:plot1}
        \addplot [draw=mycolor2, line width=1.5pt, mark=square, mark options={solid, mycolor2}, mark repeat={10}, smooth]
            table[x index=0, y index=1]{./Data/resnet50v2_gamma0.9_error_l1norm.txt};
            \label{plots:plot2}
            
   \nextgroupplot[ylabel style={align=center},
    ylabel=Middle]
        \coordinate (top) at (axis cs:0,\pgfkeysvalueof{/pgfplots/ymax});
         \addplot [draw=mycolor1, line width=1.5pt, mark=diamond, mark options={solid, mycolor1}, mark repeat={10}, smooth]
            table[x index=0, y index=2]{./Data/resnet50v2_gamma0.7_grad_l1norm.txt};
        \addplot [draw=mycolor2, line width=1.5pt, mark=square, mark options={solid, mycolor2}, mark repeat={10}, smooth]
            table[x index=0, y index=2]{./Data/resnet50v2_gamma0.7_error_l1norm.txt};

    \nextgroupplot[ylabel style={align=center},
    ylabel=Middle]
        \coordinate (top) at (axis cs:0,\pgfkeysvalueof{/pgfplots/ymax});
         \addplot [draw=mycolor1, line width=1.5pt, mark=diamond, mark options={solid, mycolor1}, mark repeat={10}, smooth]
            table[x index=0, y index=2]{./Data/resnet50v2_gamma0.9_grad_l1norm.txt};
        \addplot [draw=mycolor2, line width=1.5pt, mark=square, mark options={solid, mycolor2}, mark repeat={10}, smooth]
            table[x index=0, y index=2]{./Data/resnet50v2_gamma0.9_error_l1norm.txt};
    
    \nextgroupplot[ylabel style={align=center},
    ylabel=Lower]
        \addplot [draw=mycolor1, line width=1.5pt, mark=diamond, mark options={solid, mycolor1}, mark repeat={10}, smooth]
            table[x index=0, y index=3]{./Data/resnet50v2_gamma0.7_grad_l1norm.txt};
        \addplot [draw=mycolor2, line width=1.5pt, mark=square, mark options={solid, mycolor2}, mark repeat={10}, smooth]
            table[x index=0, y index=3]{./Data/resnet50v2_gamma0.7_error_l1norm.txt};

    \nextgroupplot[ylabel style={align=center},
    ylabel=Lower]
        \addplot [draw=mycolor1, line width=1.5pt, mark=diamond, mark options={solid, mycolor1}, mark repeat={10}, smooth]
            table[x index=0, y index=3]{./Data/resnet50v2_gamma0.9_grad_l1norm.txt};
        \addplot [draw=mycolor2, line width=1.5pt, mark=square, mark options={solid, mycolor2}, mark repeat={10}, smooth]
            table[x index=0, y index=3]{./Data/resnet50v2_gamma0.9_error_l1norm.txt};
            
    \coordinate (bot) at (axis cs:1,\pgfkeysvalueof{/pgfplots/ymin});
    \end{groupplot}

    \path (top|-current bounding box.north)--
          coordinate(legendpos)
          (bot|-current bounding box.north);
    \matrix[
        matrix of nodes,
        anchor=south,
        draw,
        inner sep=0.2em,
        draw
      ]at([yshift=1ex,xshift=-5ex]legendpos)
      {
        \ref{plots:plot1}& gradient term, $\gv_t$ &[3pt]
        \ref{plots:plot2}& memory  term, $\mv_t$ &[3pt]\\};
\end{tikzpicture}
    \vspace{-0.5cm}
    \caption{$L_1$ norm of gradient and error term for upper, middle, and lower layers from the ResNet50V2 when
    $\gamma=0.7$ (left) and $\gamma=0.9$ (right).}
    \label{fig:resnet50v2_l1norm}
    \vspace{-0.5cm}
\end{figure*}
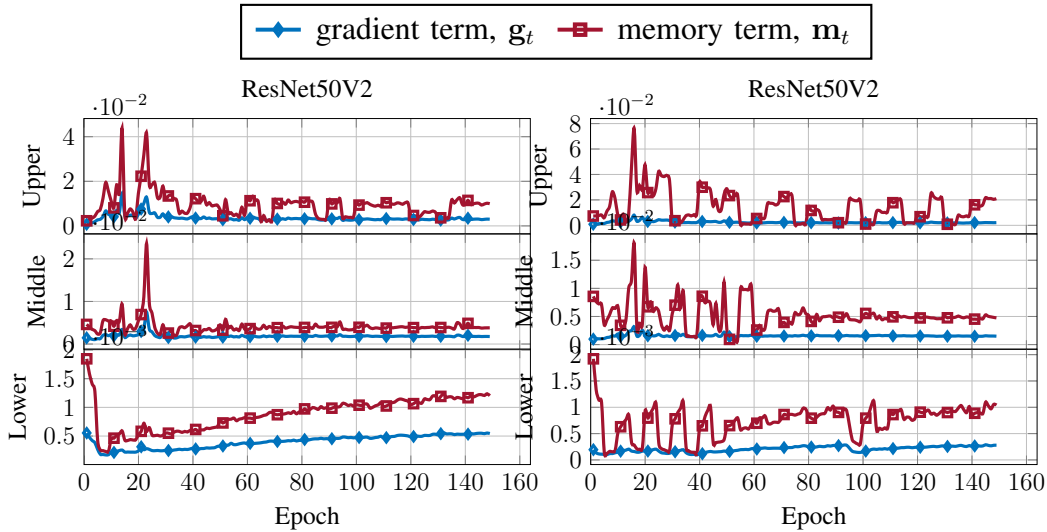

Next, in Figs. \ref{fig:resnet50v2_l1norm} and , we investigate the magnitude of the memory term, $\mv_t$, versus the gradient term, $\gv_t$, in the error feedback mechanisms with $\gamma=0.7$ and $0.9$, respectively. Note that the choice of magnitude as a performance metric has been justified in \cite{karimireddy2019error}. Here, for numerical stability, we change the scale $b_t$ only every 5 epochs and use the same $b_t$ for the next 5 epochs. Now, we clarify various aspects of $\coiii$. We begin by clarifying the simulation settings and then revisit the three main ingredients of $\coiii$ from a numerical point of view. 



%
%
We note that the differences between the magnitudes of $\mv_t$ and $\gv_t$ in these two figures are quite stable for this choice of $\gamma$. 
By comparing these two figures and our results for other $\gamma$ (not shown to avoid repetition), we conclude that $\gamma=0.7$ is a fairly good choice that leads to a very stable and small difference between the magnitudes of $\mv_t$ and $\gv_t$.
%



\newtheorem{definition}{Definition}
\newtheorem{lemma}{Lemma}
\newtheorem{theorem}{Theorem}

\section{Convergence Proof}
\label{sec:Theoritical proof}
In this section, we explore the upper bound of the optimal gap of SGD when applying $\coiii$ algorithm, which includes error feedback, fp conversion and $\gennorm$ modelling. By analyzing these bounds, we can gain insight into the convergence of SGD with $\coiii$. Our proof adapts the one in \cite{stich2018sparsified} to suit specifically for $\coiii$.  Through rigorous analysis, we demonstrate the effectiveness of SGD in minimizing the objective function and achieving convergence in the optimization process.

\subsection{Assumptions}
\begin{assumption}[Bounded true gradient]
\label{ass:Bounded gradient}
The function $ \nabla \Lcal(\wv)$ is $L_2$ bounded, that is 
\ea{
\| \nabla \Lcal(\wv_t) \|_2 \leq G.
}
\end{assumption}

\begin{assumption}[$L$-smoothness]
\label{ass:L smooth}
The function $\Lcal(\wv)$ is  $L$-smooth, that is 
\ea{
\labs \Lcal(\wv) -  \lb \Lcal(\vv) + \nabla \Lcal(\vv)^{\Tsf} (\wv- \vv) \rb  \rabs \leq  \f {L} 2 \| \wv - \vv \|.
}

\end{assumption}

\begin{assumption}[$\mu$-strongly convex]
The function $\Lcal(\wv)$ is  $\mu$ strongly convex, that is 
\ea{
\labs \Lcal(\wv) -  \lb \Lcal(\vv) + \nabla \Lcal(\vv)^{\Tsf} (\wv- \vv) \rb  \rabs \geq  \f {\mu} 2 \| \wv - \vv \|.
}

\end{assumption}

\begin{assumption}[Gen-norm assumption:]
\label{ass:Gen-norm assumption}
At iteration $t$ and for all $u \in [U]$, we have 
\ea{
g_t^{(u)} \sim \gennorm( \nabla \Lcal(\wv_t),\Dcal^{(u)}),\sgs_t,\kappa_t),
}
for some sequence of $\{\sgs_t,\kappa_t\}_{t \in [T]}$ which is non-increasing and with
$\sgs_t, \kappa_t <  2$.
%
\end{assumption}

\subsection{Quantization error bounding}
\label{subsec:Quantization error bounding}
Recall that in Section~\ref{subsec:CO}, the fp compression is represented by  $\sf[sgn, exp, mant]$ where the input will be compressed into the closest value of the form  
$
c_{\sf sgn} \cdot c_{\sf scale} \cdot c_{\sf mant}  \cdot 2^{ c_{\sf exp}},
$
where $c_{\sf sgn}\in\{+1,-1\}$, $c_{\sf mant} \in \Mc=\{1+a_1 2^{-1}+...+a_{\sf mant}2^{- \sf mant}\}$ is determined by ${\sf mant}$ bits $a_1,\ldots, a_{\sf mant}\in\mathbb{F}_2$, and $c_{\sf exp} \in \Ec =\{-(2^{\sf exp-1}-2),...,0,1,...,2^{\sf exp-1}-1\}$ can be determined by ${\sf exp}$ bits. Also, to ease the notation, in what follows, we use the notation $c_{\sf scale}=2^b$.

Next, we set $c_{\sf scale}  = (1+L)2^{-(2^{\sf exp-1}-2)}$, which leads to the boundaries of the quantizer $\pm B$ with 
\ea{
B &= (1+L)(1+2^{-1}+...+2^{- \sf mant})2^{[(2^{\sf exp-1}-1)-(2^{\sf exp-1}-2)]} \nonumber \\
&< (1+L) \cdot \lb 1+ \frac{\frac{1}{2}}{1-\frac{1}{2}} \rb \cdot 2 = 4(1+L).
\label{eq: quantizer boundary}
}
With this choice of $c_{\sf scale}$, we can bound the variance of the quantization error as follows:
\begin{lemma}\label{lem:error_bound}
For SGD with $\coiii$ with $c_{\sf scale}=(1+L)2^{-(2^{\sf exp-1}-2)}$, the variance of the quantization error $E_t\triangleq G_t - Q_{\sf fp}(G_t)$ is bounded as      
\ea{
\Ebb \lsb E_t^2 \rsb \leq  2^{4-2 \sf mant} + 0.3.
}
\end{lemma}
\begin{IEEEproof}
Define $\Delta = c_{\sf scale} 2^{-\sf mant} \cdot 2^{2^{\sf exp-1}-1}$ the largest interval of non-uniform quantization levels of $Q_{fp}$. We then bound the variance of the quantization error arising from floating point as follows:
\ea{
&\Ebb [ E_t^2  ]  = \Ebb[ E_t^2  |  | G_t|  \leq B] +   \Ebb[ E_t^2  |  | G_t| >B]  \nonumber \\
& \overset{(a)}{\leq} \f {\Delta^2} 4 + \Ebb[ E_t^2  | | G_t|>B] \overset{(b)}{=} \f {\Delta^2} 4 + 2\Ebb[ E_t^2  |  G_t>B], \label{eqn:error_var_bound}
}
where in (a) we have used the fact that the expected value of quantization error for $G_t<B$ can be upper bounded by $\Delta/2$, implying that the variance is bounded by $\Delta^2/4$; (b) is due the symmetry of $G_t$ and $E_t$. 
Furthermore, we have used the fact that the second moment of a distribution with support $[-1/2,1/2]$ with the maximum variance is the equiprobable distribution on $\{1/2,1/2\}$.
%
%
Now, with the $\gennorm$ assumption, the tail function for the $\gennorm$ for this interval is decreasing in $B$ and increasing in $\be$, so that one can  obtain the upper bound
$
\Ebb [ E_t^2  |  G_t>B  ]  \leq  0.15,
\label{eq:bound tail 2}
$
with a maximum at $\be=1$ by numerically integrating $\gennorm(x,\mu, \al, \be) = \gennorm(x, 0, 1, [0,2])$. Note that in the above upper bound, we set $L=0$ because we need an upper bound for the quantization error, and any other value of $L>0$ will result in a smaller tail that falls outside the bound. 
Combining the definition of $\Delta$, the bound on the conditional variance, and \eqref{eqn:error_var_bound} shows the result.
\end{IEEEproof}

\subsection{Convergence analysis}
Having established an upper bound on the variance of the quantization error, we are now ready to show the convergence of the the SGD with the proposed $\coiii$. Following the perturbed iterate analysis of \cite{stich2018sparsified}, we first define a virtual sequence $\{\Tilde{\wv}_t\}_{t\geq 0}$ as follows,
\ea{\Tilde{\wv}_0 = \Hat{\wv}_0,  \Tilde{\wv}_{t+1} = \Tilde{\wv}_t - \eta \gv_t, \gv_t = \nabla\Lcal\lb\Hat{\wv}_t,\xv_{t,i}\rb,
    \label{eq: virtual sequence definition}}
where the sequences $\{\Hat{\wv}_t\}_{t>0}$ is the same as in $\coiii$ algorithm. Notice that
\ea{\Tilde{\wv}_t - \Hat{\wv}_t=(\wv_0-\sum_{j=0}^{t-1}\eta\gv_j) - (\wv_0-\sum_{j=0}^{t-1}\eta \Hat{\gv}_j)=\eta\mv_t.} The following lemma bounds the distance between the virtual sequence and the optimal weight.

\begin{lemma}{\cite[Lemma 3.1]{stich2018sparsified}}
    Let $\{\Hat{\wv}_t\}_{t\geq0}$ and $\{\Tilde{\wv}_t\}_{t\geq0}$ be defined as in $\coiii$ algorithm and \eqref{eq: virtual sequence definition}, respectively. Then
    \begin{align}
    \Ebb {\| \Tilde{\wv}_{t+1} - \wv^*\|}^2 &\leq (1-\frac{\eta\mu}{2})\Ebb {\| \Tilde{\wv}_{t} - \wv^*\|}^2 + {\eta}^2G^2\nonumber \\
    &-\eta e_t+\eta^3 (\mu+2L) \Ebb\| \mv_t \|^2, 
    \end{align}
    where $\wv^*:= {arg\,min}_{\wv \in \Rbb^d}\Lcal(\wv)$ and $e_t:=\Ebb \Lcal(\hat{\wv}_t)-\Lcal^*$
\end{lemma}
The memory term $\mv_t$ is equal to the quantization error $E_t$ when applying $\coiii$ algorithm, so the upper bound in Lemma~\ref{lem:error_bound} becomes handy later. Before we prove the convergence, we show the following lemma that is a modified version of \cite[Lemma 3.3]{stich2018sparsified} without the shift parameter $w_t$:
\begin{lemma}{\cite[Lemma 3.3]{stich2018sparsified}}
    Let $\{a_t\}_{t\geq0}, a_t\geq 0, \{e_t\}_{t\geq0}, e_t\geq 0$, be sequences satisfying
    \ea{a_{t+1}\leq(1-\frac{\mu\eta}{2})a_t+{\eta}^2 M+\eta^3 N-\eta e_t, \label{eq: lemma 2 eq 1}}
    for constants $M,N\geq0, \mu\geq0$. Then
    \ea{\frac{1}{T}\sum_{t=0}^{T-1}e_t \leq \frac{1}{T\eta}a_0+\eta M+\eta^2 N.}
\end{lemma}
\begin{IEEEproof}
    Multiplying \eqref{eq: lemma 2 eq 1} with $\frac{1}{\eta}$ yields
    \ea{\frac{1}{\eta}a_{t+1}\leq(1-\frac{\mu\eta}{2})\frac{1}{\eta}a_t+\eta M+\eta^2 N- e_t,}
    which leads to 
    \ea{\frac{1}{\eta}a_{t+1}\leq \frac{1}{\eta}a_t+\eta M + \eta^2 N- e_t,}
    because $\mu,\eta > 0$, $(1-\frac{\mu\eta}{2})a_t < a_t$. By recursively substituting $\frac{1}{\eta}a_t$, we obtain
    \ea{\frac{1}{\eta}a_{T}\leq \frac{1}{\eta}a_0+T\eta M+T\eta^2 N- \sum_{t=0}^{T-1}e_t, }
    which completes the proof.
\end{IEEEproof}

Now, following the steps in \cite[Theorem 2.4]{stich2018sparsified}, we put everything together and prove the convergence of SGD with the proposed $\coiii$ in what follows.
\begin{theorem} 
    Let $\Lcal$ be L-smooth, $\mu$-strongly convex, $\| \nabla \Lcal(\wv_t) \|_2 \leq G$ for $t = 0,..., T-1$, where $\{\wv_t\}_{t\geq0}$ are generated according to $\coiii$ algorithm for stepsizes $\eta=\frac{1}{\sqrt{T}}$. Then,
    \begin{align}
    \Ebb \Lcal(\Bar{\wv}_T)-\Lcal^* &\leq \frac{1}{\sqrt{T}}{\|\Hat{\wv}_0 - \wv^*\|}^2+\frac{1}{\sqrt{T}}G^2\nonumber \\
    &+\frac{1}{T}(\mu+2L)(2^{4-2\sf mant} + 0.3),
    \label{eq:convergence bound}
    \end{align}
    where $\Bar{\wv}_T=\frac{1}{T}\sum_{t=0}^{T-1}\Hat{\wv}_t$.
\end{theorem}
The proof uses the fact that $\Lcal$ is convex, i.e. $\Ebb \Lcal(\Bar{\wv}_T)-\Lcal^* \leq \frac{1}{T}\sum_{t=0}^{T-1}e_t$ with combining the results of Lemma 2 and 3, for constants $M=G^2$ and $N=(\mu+2L)(2^{4-2\sf mant} + 0.3)$.


\section{Simulation}\label{sec:simulation}
We conclude the paper with some plots regarding the overall performance of the proposed $\coiii$ with two compared compression schemes: Tinyscript\cite{fangcheng2020tinyscript} and Count Sketch\cite{rothchild2020countsketch}.
The parameters are as shown in Tab. \ref{tab:DNN parameters} and \ref{tab:total network parameters}.

In Fig. \ref{fig:accuracy_resnet50v2} left, we plot the test accuracy against the epoch for the ResNet50V2. 
For $\coiii$, consider the fp4 quantization with $[\sf sgn \ \sf mant \ \sf exp]=[1, \ 2, \ 1]$ and with $\gamma=0,0.7$. 
 %
 %
Note that for $\gamma = 0$, it is essentially not using error feedback while $\gamma = 1$ is using a non-decaying error feedback.
Through our experimentation, we observe that as predicted in our results in Sec.~\ref{subsec:error_mag}, with $\gamma=0.7$, the proposed $\coiii$ can provide a reasonably good accuracy performance that is comparable to the other two compression schemes and even full SGD computation, while requiring $\Rsf_{fp4,Res}=2.46\times 10^{12}$ bits, a significantly less communication resource.
Moreover, similar results for fp8 with $[\sf sgn \ \sf mant \ \sf exp]=[1, \ 5, \ 2]$ are presented in Fig.~\ref{fig:accuracy_resnet50v2} right, where $\gamma=0.7$ seems to be a reasonable choice. In regards to the jumps observed in the purple curve ($\coiii$ with fp8 training using error feedback), we utilize the same parameter of the $\gennorm$ distribution for a duration of 5 epochs rather than for each iteration. This is because distribution fitting can be time-consuming and redundant. The distribution fitting parameter is utilized for the constant $b$ mentioned in Sec. \ref{subsec:CO_DNN}. However, if the distribution changes during this duration, significant jumps in accuracy may occur.
We repeat the simulation for NASNetMobile in Fig.~\ref{fig:accuracy_nasnetmobile_8x}, which shows similar results that with $\Rsf_{fp4,NAS}=3.43\times10^{11}$ bits, the proposed $\coiii$ with $\gamma=0.9$ is able to achieve the accuracy performance that is comparable to the full SGD computation. We note that the choice $\gamma=0.9$ can also be justified by looking into our results for error magnitudes (not shown to save space). 

Finally, we plot the test accuracy against $\sf R$ the communication overhead for NASNetMobile in Fig.~\ref{fig:overhead_acc}.
In this figure, five schemes are considered, namely the proposed $\coiii$ with fp4, the proposed $\coiii$ with fp8, the scheme that employs fp8 followed by the $\topK$, Count Sketch with the 8x compression rate, and the Tinyscript with 4 bits.
Moreover, for the $\topK$, we adopt $k=50\%$ such that only half of the entries in the gradients are retained.
Each scheme is trained for 150 epochs, and the accuracy is tested on the already trained model.
One observes in Fig.~\ref{fig:overhead_acc} that the proposed scheme is capable of achieving an accuracy that is comparable to or even better than the other compression schemes while using significantly less communication overhead.
\newenvironment{customlegend}[1][]{%
    \begingroup
    \csname pgfplots@init@cleared@structures\endcsname
    \pgfplotsset{#1}%
}{%
    \csname pgfplots@createlegend\endcsname
    \endgroup
}%

\def\addlegendimage{\csname pgfplots@addlegendimage\endcsname}

\begin{figure}
    \centering
	\begin{tikzpicture}[scale=0.08]
    \definecolor{mycolor1}{rgb}{0.00000,0.44706,0.74118}%
    \definecolor{mycolor2}{rgb}{0.63529,0.07843,0.18431}%
    \definecolor{mycolor3}{rgb}{0.00000,0.49804,0.00000}%
    \definecolor{mycolor4}{rgb}{0.60000,0.19608,0.80000}%
    \definecolor{mycolor5}{rgb}{0.00000,0.00000,0.50196}%
    \definecolor{mycolor6}{rgb}{1.00000,0.84314,0.00000}%
    \begin{customlegend}[legend entries={Count Sketch,TinyScript (with EF),
                                        $\coiii$ (without EF),$\coiii$ (with EF),
                                        Uncompressed}]
    \addlegendimage{draw=mycolor1, line width=1.5pt, mark=diamond, mark options={solid, mycolor1}}
    \addlegendimage{draw=mycolor2, line width=1.5pt, mark=square, mark options={solid, mycolor2}}
    \addlegendimage{draw=mycolor3, line width=1.5pt, mark=square, mark options={solid, mycolor3}}
    \addlegendimage{draw=mycolor4, line width=1.5pt, mark=diamond, mark options={solid, mycolor4}}
    \addlegendimage{draw=mycolor5, line width=1.5pt, mark=square, mark options={solid, mycolor5}}
    \end{customlegend}
\end{tikzpicture}
    \vspace{-0.2cm}
    \caption{Shared legend for Figs. \ref{fig:accuracy_resnet50v2} and \ref{fig:accuracy_nasnetmobile_8x}.}
    \label{fig:accuracy_legend}
    \vspace{-0.5cm}
\end{figure}
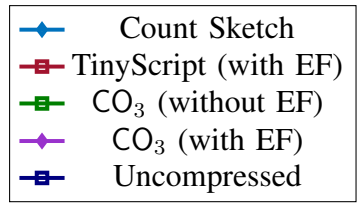



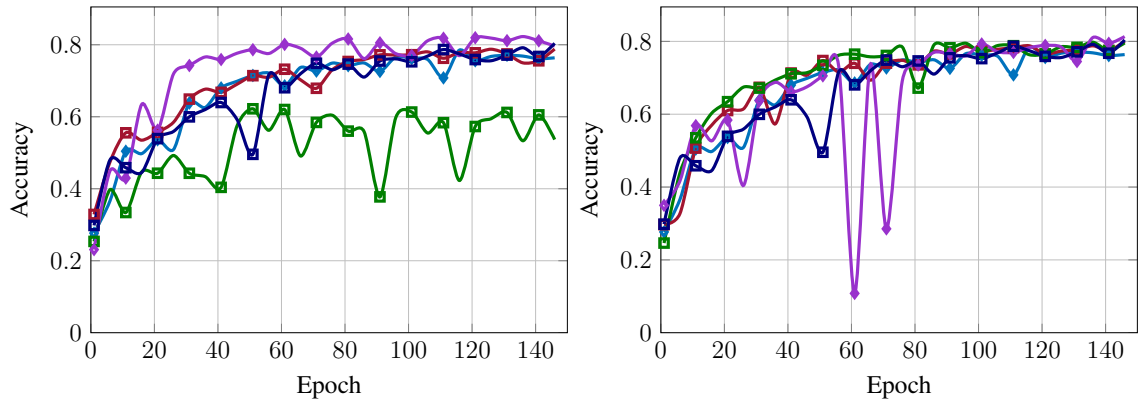
\begin{figure} 
    \centering
    \begin{subfigure}{.45\textwidth}
        \centering
    	\begin{tikzpicture}[scale = 0.8]
    \definecolor{mycolor1}{rgb}{0.00000,0.44706,0.74118}%
    \definecolor{mycolor2}{rgb}{0.63529,0.07843,0.18431}%
    \definecolor{mycolor3}{rgb}{0.00000,0.49804,0.00000}%
    \definecolor{mycolor4}{rgb}{0.60000,0.19608,0.80000}%
    \definecolor{mycolor5}{rgb}{0.00000,0.00000,0.50196}%
    \definecolor{mycolor6}{rgb}{1.00000,0.84314,0.00000}%
    \begin{axis}[
    ymin=0,
    xmin=0,
    xmax=150,
    xlabel={Epoch},
    ylabel={Accuracy},
    grid=both,
    height=7cm, 
    width=9.5cm,
    ]
    \coordinate (top) at (axis cs:1,\pgfkeysvalueof{/pgfplots/ymax});
        \addplot [draw=mycolor1, line width=1.5pt, mark=diamond, mark options={solid, mycolor1}, mark repeat={2}, smooth, each nth point=5, filter discard warning=false, unbounded coords=discard ]
            table[x index=0, y index=1]{./Data/resnet50v2_acc_8x.txt};
            \label{plots:plot1_acc_8x}
        \addplot [draw=mycolor2, line width=1.5pt, mark=square, mark options={solid, mycolor2}, mark repeat={10}, smooth, mark repeat={2}, smooth, each nth point=5, filter discard warning=false, unbounded coords=discard ]
            table[x index=0, y index=2]{./Data/resnet50v2_acc_8x.txt};
            \label{plots:plot2_acc_8x}
        \addplot [draw=mycolor3, line width=1.5pt, mark=square, mark options={solid, mycolor3}, mark repeat={10}, smooth, mark repeat={2}, smooth, each nth point=5, filter discard warning=false, unbounded coords=discard ]
            table[x index=0, y index=3]{./Data/resnet50v2_acc_8x.txt};
            \label{plots:plot3_acc_8x}
        \addplot [draw=mycolor4, line width=1.5pt, mark=diamond, mark options={solid, mycolor4}, mark repeat={10}, smooth, mark repeat={2}, smooth, each nth point=5, filter discard warning=false, unbounded coords=discard ]
            table[x index=0, y index=4]{./Data/resnet50v2_acc_8x.txt};
            \label{plots:plot4_acc_8x}
        \addplot [draw=mycolor5, line width=1.5pt, mark=square, mark options={solid, mycolor5}, mark repeat={10}, smooth, mark repeat={2}, smooth, each nth point=5, filter discard warning=false, unbounded coords=discard ]
            table[x index=0, y index=5]{./Data/resnet50v2_acc_8x.txt};
            \label{plots:plot5_acc_8x}
    \coordinate (bot) at (axis cs:1,\pgfkeysvalueof{/pgfplots/ymin});
    \end{axis}
    /%
\end{tikzpicture}
        \vspace{-1.0cm}
        \label{fig:accuracy_resnet50V2_8x}
    \end{subfigure}
    \begin{subfigure}{.45\textwidth}    
        \centering
    	\begin{tikzpicture}[scale = 0.8]
    \definecolor{mycolor1}{rgb}{0.00000,0.44706,0.74118}%
    \definecolor{mycolor2}{rgb}{0.63529,0.07843,0.18431}%
    \definecolor{mycolor3}{rgb}{0.00000,0.49804,0.00000}%
    \definecolor{mycolor4}{rgb}{0.60000,0.19608,0.80000}%
    \definecolor{mycolor5}{rgb}{0.00000,0.00000,0.50196}%
    \definecolor{mycolor6}{rgb}{1.00000,0.84314,0.00000}%
    \begin{axis}[
    ymin=0,
    xmin=0,
    xmax=150,
    xlabel={Epoch},
    ylabel={Accuracy},
    grid=both,
    height=7cm,
    width=9.5cm,
    ]
    \coordinate (top) at (axis cs:1,\pgfkeysvalueof{/pgfplots/ymax});
        \addplot [draw=mycolor1, line width=1.5pt, mark=diamond, mark options={solid, mycolor1}, mark repeat={2}, smooth, each nth point=5, filter discard warning=false, unbounded coords=discard ]
            table[x index=0, y index=1]{./Data/resnet50v2_acc_4x.txt};
            \label{plots:plot1_acc_8x}
        \addplot [draw=mycolor2, line width=1.5pt, mark=square, mark options={solid, mycolor2}, mark repeat={10}, smooth, mark repeat={2}, smooth, each nth point=5, filter discard warning=false, unbounded coords=discard ]
            table[x index=0, y index=2]{./Data/resnet50v2_acc_4x.txt};
            \label{plots:plot2_acc_8x}
        \addplot [draw=mycolor3, line width=1.5pt, mark=square, mark options={solid, mycolor3}, mark repeat={10}, smooth, mark repeat={2}, smooth, each nth point=5, filter discard warning=false, unbounded coords=discard ]
            table[x index=0, y index=3]{./Data/resnet50v2_acc_4x.txt};
            \label{plots:plot3_acc_8x}
        \addplot [draw=mycolor4, line width=1.5pt, mark=diamond, mark options={solid, mycolor4}, mark repeat={10}, smooth, mark repeat={2}, smooth, each nth point=5, filter discard warning=false, unbounded coords=discard ]
            table[x index=0, y index=4]{./Data/resnet50v2_acc_4x.txt};
            \label{plots:plot4_acc_8x}
        \addplot [draw=mycolor5, line width=1.5pt, mark=square, mark options={solid, mycolor5}, mark repeat={10}, smooth, mark repeat={2}, smooth, each nth point=5, filter discard warning=false, unbounded coords=discard ]
            table[x index=0, y index=5]{./Data/resnet50v2_acc_4x.txt};
            \label{plots:plot5_acc_8x}
    \coordinate (bot) at (axis cs:1,\pgfkeysvalueof{/pgfplots/ymin});
    \end{axis}
    /%
\end{tikzpicture}
        \vspace{-1.0cm}
        \label{fig:accuracy_resnet50v2_4x}
    \end{subfigure}
    \vspace{-0.3cm}
    \caption{Test accuracy of ResNet50V2 with fp4 (left) and fp8 (right) training with  communication overhead of $\Rsf_{fp4,Res} = \num{2.46e12}$ and  $\Rsf_{fp8,Res} = \num{3.24e12}$ bits.}
    \label{fig:accuracy_resnet50v2}
    \vspace{-0.5cm}
\end{figure}



\begin{figure} 
    \centering
    \begin{minipage}{.45\textwidth}
        \centering
    	\begin{tikzpicture}[scale = 0.8]
    \definecolor{mycolor1}{rgb}{0.00000,0.44706,0.74118}%
    \definecolor{mycolor2}{rgb}{0.63529,0.07843,0.18431}%
    \definecolor{mycolor3}{rgb}{0.00000,0.49804,0.00000}%
    \definecolor{mycolor4}{rgb}{0.60000,0.19608,0.80000}%
    \definecolor{mycolor5}{rgb}{0.00000,0.00000,0.50196}%
    \definecolor{mycolor6}{rgb}{1.00000,0.84314,0.00000}%
    \begin{axis}[
    ymin=0,
    xmin=0,
    xmax=150,
    xlabel={Epoch},
    ylabel={Accuracy},
    grid=both,
    height=7cm, 
    width=9.5cm,
    ]
    \coordinate (top) at (axis cs:1,\pgfkeysvalueof{/pgfplots/ymax});
        \addplot [draw=mycolor1, line width=1.5pt, mark=diamond, mark options={solid, mycolor1}, mark repeat={2}, smooth, each nth point=5, filter discard warning=false, unbounded coords=discard ]
            table[x index=0, y index=1]{./Data/nasnetmobile_acc_8x.txt};
            \label{plots:plot1_acc_8x}
        \addplot [draw=mycolor2, line width=1.5pt, mark=square, mark options={solid, mycolor2}, mark repeat={10}, smooth, mark repeat={2}, smooth, each nth point=5, filter discard warning=false, unbounded coords=discard ]
            table[x index=0, y index=2]{./Data/nasnetmobile_acc_8x.txt};
            \label{plots:plot2_acc_8x}
        \addplot [draw=mycolor3, line width=1.5pt, mark=square, mark options={solid, mycolor3}, mark repeat={10}, smooth, mark repeat={2}, smooth, each nth point=5, filter discard warning=false, unbounded coords=discard ]
            table[x index=0, y index=3]{./Data/nasnetmobile_acc_8x.txt};
            \label{plots:plot3_acc_8x}
        \addplot [draw=mycolor4, line width=1.5pt, mark=diamond, mark options={solid, mycolor4}, mark repeat={10}, smooth, mark repeat={2}, smooth, each nth point=5, filter discard warning=false, unbounded coords=discard ]
            table[x index=0, y index=4]{./Data/nasnetmobile_acc_8x.txt};
            \label{plots:plot4_acc_8x}
        \addplot [draw=mycolor5, line width=1.5pt, mark=square, mark options={solid, mycolor5}, mark repeat={10}, smooth, mark repeat={2}, smooth, each nth point=5, filter discard warning=false, unbounded coords=discard ]
            table[x index=0, y index=5]{./Data/nasnetmobile_acc_8x.txt};
            \label{plots:plot5_acc_8x}
    \coordinate (bot) at (axis cs:1,\pgfkeysvalueof{/pgfplots/ymin});
    \end{axis}
    /%
\end{tikzpicture}
        \vspace{-0.8cm}
        \caption{Test accuracy of NASNetMobile with fp4 training. The communication overhead is $\Rsf_{fp4,NAS} = \num{3.43e11}$ bits.\\}
        \label{fig:accuracy_nasnetmobile_8x}
    \end{minipage}
    \hspace{0.3cm}
    \begin{minipage}{.45\textwidth}    
        \centering
    	 \begin{tikzpicture}[scale = 0.75]
    \definecolor{mycolor1}{rgb}{0.00000,0.44706,0.74118}%
    \definecolor{mycolor2}{rgb}{0.63529,0.07843,0.18431}%
    \definecolor{mycolor3}{rgb}{0.00000,0.49804,0.00000}%
    \definecolor{mycolor4}{rgb}{0.60000,0.19608,0.80000}%
    \definecolor{mycolor5}{rgb}{0.00000,0.00000,0.50196}%
    \definecolor{mycolor6}{rgb}{1.00000,0.84314,0.00000}%
    \begin{semilogxaxis}[   
        xlabel={$\Rsf$ (bits)},
        ylabel={Test accuracy},
        ymin=0.3,
        ymax=1,
        xmin=0,
        legend style={
        at={(0.5,-0.6)},
        anchor=north,
        legend columns=1
        },
        grid=both,
        height=4 cm,
        width= 8 cm ]
        \addplot[
        scatter,only marks,scatter src=explicit symbolic,
        scatter/classes={
            a={mark=*,mycolor1},
            b={mark=*,mycolor2},
            c={mark=*,mycolor3},
            d={mark=*,mycolor4},
            e={mark=*,mycolor5}
        }
    ]
    table[x=x,y=y,meta=label]{
        x    y    label
        3.43e11 7.477999925613403320e-01 a
        3.65e11 7.918999791145324707e-01 b
        1.99e12  7.471e-01 c
        1.99e12  4.007e-01 d
        1.99e12  6.272e-01 e
    };
    \legend{$\coiii$ (fp$4$ with EF $(\gamma=0.9)$),
            $\coiii$ (fp$8$ with EF $(\gamma=0.7)$),
            fp$8$ + $\topK$ ($ k = 50\%$),
            Count Sketch (8x),
            Tinyscript (4 bits with EF $(\gamma=0.7)$)}
\end{semilogxaxis}
\end{tikzpicture}
        \caption{The communication overhead versus test accuracy of different compression schemes of NASNetMobile. The uncompressed scheme requires $\mathsf{R}=\num{1.59e13}$ bits.}
        \label{fig:overhead_acc}
    \end{minipage}
    \vspace{-0.5cm}
\end{figure}


\section{Conclusion}\label{sec:conclude}
In this paper, we have investigated the problem of communication-efficient distributed DNN training. A novel framework, called $\coiii$, has been proposed for the federated training of a centralized model. 
When specialized to DNN models, gradients are first quantized via floating-point conversion, lossless compressed by Huffman codes, and then corrected by adding the quantization error to the gradients in the next epoch. %
The effectiveness of  $\coiii$ hinges on the fact that the distribution over the gradients can be well-defined. 
For the DNN case,  we argue that the floating point conversions and Huffman coding can be efficiently designed by assuming that the gradients are i.i.d. distributed in each layer and are independent across layers and epochs (but with possibly different parameters). 
We refer to this assumption as the $\gennorm$ assumption. 
In this paper, this assumption has been verified by a series of simulations, which has in turn allowed us to build a statistical model for gradients in DNN training.
The theoretical proof of quantized SGD convergence with $\gennorm$ assumption and error feedback is also provided to further enhance the feasibility of proposed approach.
Extensive simulations have been conducted to demonstrate that $\coiii$ provides  competitive DNN training performance  at a significantly lower communication overhead. In particular,   $\coiii$ is capable of achieving test accuracy performance that is comparable to the network trained with full SGD computation.

\bibliographystyle{IEEEtran}
\bibliography{IEEEabrv,bib_JSAC}
     
\end{document}